\documentclass[manuscript,screen,acmlarge]{acmart} 
\usepackage{pifont} 
\usepackage{xcolor}
\usepackage{xurl}
\usepackage{booktabs}
\usepackage[ruled]{algorithm2e}
\usepackage{algpseudocode}
\usepackage{enumerate}
\usepackage[english]{babel}
\usepackage{blindtext}
\usepackage[caption=false,font=footnotesize,subrefformat=parens]{subfig}
\usepackage{amsmath}

\usepackage{amssymb}
\usepackage{xspace}
\usepackage{multirow}
\usepackage{threeparttable}
\usepackage{float}
\usepackage{flafter}
\usepackage{comment}
\usepackage[skip=12pt]{caption}
\usepackage{graphicx}
\usepackage{amssymb}
\usepackage{pifont}
\usepackage{enumitem}

\def\fig{Fig.\xspace}

\def\sec{Sec.\xspace}
\def\tab{Tab.\xspace}

\def\ie{{\textit{i.e.}\xspace}} 
\def\eg{{\textit{e.g.}\xspace}}
 
\def\etal{{\textit{et al.}\xspace}}
\def\etc{{\textit{etc}\xspace}} 

\newcommand{\head}[1]{{\noindent \textbf{#1:}}}

\graphicspath{{./figures/}}
\graphicspath{{./pdf/}}
\DeclareGraphicsExtensions{.pdf,.jpeg,.png}

\ifodd 1
\newcommand{\com}[1]{\textbf{\color{red}(COMMENT: #1)}} 
\newcommand{\todo}[1]{\textbf{{\color{orange}(TODO: #1)}}}
\else

\newcommand{\com}[1]{}
\newcommand{\todo}[1]{}
\fi

\renewcommand\footnotetextcopyrightpermission[1]{} 
\setcopyright{none}

\acmDOI{}

\acmISBN{}

\acmConference[Arxiv]{}
\acmYear{2026}
\copyrightyear{}
\acmSubmissionID{}

\acmPrice{}

\def\sysname{\textsc{TaFall}\xspace}
\begin{document}

\title[\sysname]{\sysname: Balance-Informed Fall Detection via Passive Thermal Sensing}

\begin{anonsuppress}
	\author{Chengxiao Li}
	\email{chengxiaoli@connect.hku.hk}
	\affiliation{%
		\institution{University of Hong Kong}
		\city{Hong Kong}
		\country{China}
	}

    \author{Xie Zhang}
	\email{zhangxie@connect.hku.hk}
	\affiliation{%
		\institution{University of Hong Kong}
		\city{Hong Kong}
		\country{China}
	}

    \author{Wei Zhu}
	\email{zhuwei@wchscu.edu.cn}
	\affiliation{%
		\institution{West China Hospital, Sichuan University}
		\city{Chengdu}
		\country{China}
	}
    
    \author{Yan Jiang}
	\email{hxhljy2018@163.com}
	\affiliation{%
		\institution{West China Hospital, Sichuan University}
		\city{Chengdu}
		\country{China}
	}

	\author{Chenshu Wu}
	\email{chenshu@cs.hku.hk}
	\affiliation{%
		\institution{University of Hong Kong}
		\city{Hong Kong}
		\country{China}
	}
\end{anonsuppress}

\renewcommand{\shortauthors}{\sysname}

\begin{abstract}

Falls are a major cause of injury and mortality among older adults, yet most incidents occur in private indoor environments where monitoring must balance effectiveness with privacy. Existing privacy-preserving fall detection approaches, particularly those based on radio frequency sensing, often rely on coarse motion cues, which limits reliability in real-world deployments. We introduce \sysname, a balance-informed fall detection system based on low-cost, privacy-preserving thermal array sensing. 
The key insight is that \sysname models a fall as a process of balance degradation and detects falls by estimating pose-driven biomechanical balance dynamics. 
To enable this capability from low-resolution thermal array maps, we propose (i) an appearance-motion fusion model for robust pose reconstruction, (ii) physically grounded balance-aware learning, and (iii) pose-bridged pretraining to improve robustness. \sysname achieves a detection rate of 98.26\% with a false alarm rate of 0.65\% on our dataset with over 3,000 fall instances from 35 participants across diverse indoor environments. In 27 day deployments across four homes, \sysname attains an ultra-low false alarm rate of 0.00126\% and a pilot bathroom study confirms robustness under moisture and thermal interference. Together, these results establish \sysname as a reliable and privacy-preserving approach to fall detection in everyday living environments.

\end{abstract}

\maketitle


\section{Introduction}
\label{sec:intro}
\begin{figure}[t]
    \centering
    \includegraphics[width=0.99\linewidth]{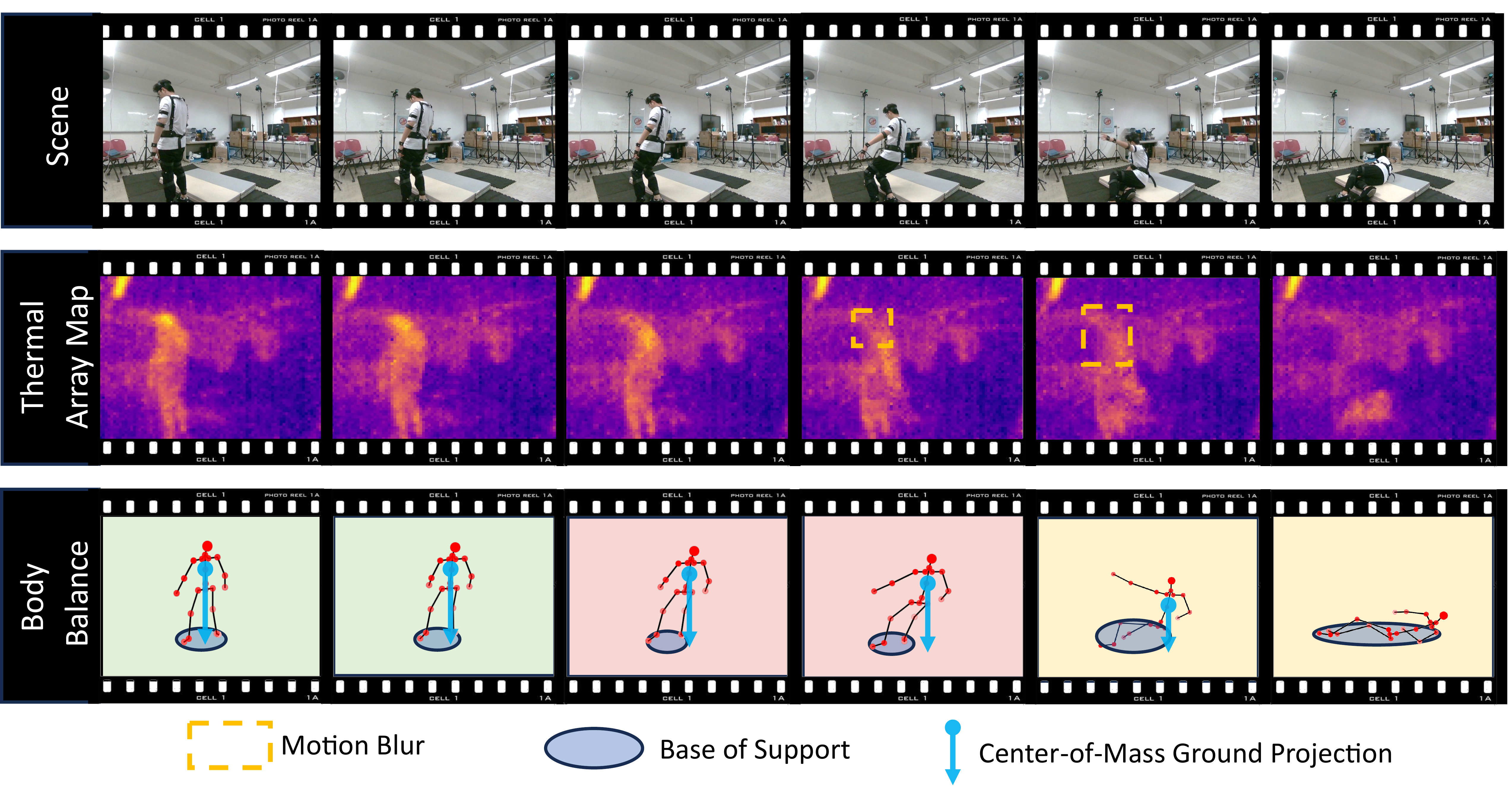}
    \vspace{-0.2in}
        \caption{Illustration of thermal array temperature maps and the corresponding body pose–based balance representation during a fall event. \rm{The yellow box highlights motion blur in the thermal temperature map induced by rapid movement. In the bottom row, the background color of each frame denotes a distinct balance state—stable balance (SB), loss of balance (LoB), and ground impact stage (GIS)—as detailed in \sec\ref{sec:design}.}}
    \label{fig:balance_fall_overview}
\end{figure}

Falls constitute a major threat to longevity and independent living among older adults, affecting 28–35\% of individuals aged 65 years or above each year \cite{world2008global}. 
Among these incidents, 20–30\% result in mild to severe injuries \cite{Lord_Sherrington_Menz_Close_2007}, accounting for approximately 10–15\% of all emergency department visits \cite{scuffham2003incidence}. 
Beyond physical harm, falls often trigger a persistent fear of falling, leading 15–55\% of older adults to restrict daily activities \cite{chung2009posttraumatic}, which in turn accelerates functional decline and degrades quality of life.

Motivated by these risks, fall detection has attracted increasing research attention in recent years. 
Existing approaches can be broadly grouped into two categories. 
The first category is wearable-based systems \cite{picerno2021wearable, hu2016pre, nait2018deep, palmerini2020accelerometer, bagala2012evaluation, bourke2016fall}, which can achieve reliable detection performance while preserving privacy. 
However, their practical adoption is often limited by user adherence issues, including charging requirements, wearing discomfort, and frequent forgetfulness.
The second category is non-contact systems that monitor falls without requiring on body devices \cite{inturi2023novel, raza2023logrf, romaissa2020fall, palipana2018falldefi, wang2016rt, wang2016wifall}. 
Among them, camera-based approaches \cite{inturi2023novel, raza2023logrf, romaissa2020fall} can deliver strong accuracy, but raise substantial privacy concerns, especially in sensitive indoor spaces such as bedrooms and bathrooms where many falls occur \cite{moreland2021descriptive}. 
Such concerns significantly reduce user acceptance in real deployments \cite{mujirishvili2023acceptance}. 
To address these privacy issues, radio frequency (RF) sensing has recently emerged as a promising alternative \cite{hu2021defall, zhang2023lt, ji2022sifall}. 
RF-based systems enable device-free and non-visual monitoring, offering a more privacy-preserving solution.

Despite their promise, existing RF-based fall detection approaches suffer from a fundamental limitation: \emph{their spatial resolution is insufficient to resolve fine-grained human body dynamics}, resulting in elevated false alarm rates in real-world deployment.
Most prior systems rely on coarse motion signatures, such as abrupt height changes \cite{zhang2023lt} or velocity spikes \cite{meng2025gr}. 
These cues are easily confounded by fall-like daily activities (\eg, sitting down quickly or bending to pick up objects) and by fast-moving non-human entities such as pets or falling items \cite{hu2024radar}. 
In practice, even a seemingly modest false alarm rate of 2\% (\eg, 1.8\% reported in controlled evaluations \cite{ji2022sifall}) per one-minute monitoring window can accumulate into dozens of false alerts per day, overwhelming caregivers and eroding user trust—an outcome that critically undermines long-term deployment.

To overcome these limitations, we revisit the fundamental biomechanical mechanism underlying a fall: \emph{the failure to maintain body balance}.
According to the World Health Organization, a fall occurs when an individual inadvertently comes to rest on a lower surface, typically following a transient yet decisive \textbf{loss of balance} \cite{world2008global}, as illustrated in \fig\ref{fig:balance_fall_overview}. 
Crucially, this loss of balance phase represents a distinct physical state in which the body’s posture exceeds its stability limits, rendering the subsequent descent inevitable.
Therefore, reliable fall detection requires capturing body pose dynamics that govern balance states, rather than relying on indirect and coarse indicators such as height changes or velocity fluctuations commonly adopted in existing systems \cite{zhang2023lt, meng2025gr}.

Building on this insight, we present \sysname, a balance-informed fall detection system using a low-cost thermal array sensor.
As shown in \fig\ref{fig:balance_fall_overview}, thermal arrays passively measure thermal radiation emitted by surrounding objects, including the human body, producing low-resolution temperature maps that inherently preserve privacy while retaining sufficient spatial structure to support pose dynamics estimation.
Unlike camera-based systems, thermal arrays do not capture appearance-level details and operate robustly under darkness or low-light conditions typical of bedrooms and bathrooms \cite{zhang2024tadar}.
Compared with RF modalities such as mmWave radar or WiFi sensing, thermal arrays provide finer spatial resolution\footnote{The thermal array sensor used in our prototype has a resolution of $62 \times 80$ within its field of view (FoV) $90^{\circ}\times67^{\circ}$, corresponding to angular resolutions of $(1.13^{\circ}, 1.08^{\circ})$ in azimuth and elevation, whereas a commonly used mmWave radar (TI IWR1843) provides $(15^{\circ},60^{\circ})$ resolution.} 
and stronger human–background contrast due to the consistent human body temperature, thereby enabling more accurate reconstruction of pose dynamics that are essential for balance estimation.
Leveraging these advantages, \sysname models a fall as a \emph{temporal progression of balance degradation}.
However, translating this idea into a practical fall detection system introduces significant challenges.

\noindent\textbf{\ding{182} Robust pose estimation.}
Thermal array sensors generate low-resolution (\eg, $62 \times 80$ for the sensor used in this paper), textureless temperature maps, as shown in \fig\ref{fig:balance_fall_overview}, making the recovery of fine-grained skeletal structures substantially more challenging than in RGB imagery.
Moreover, rapid movements during a fall induce motion blur, which further obscures body contours and disrupts temporal consistency.
These sensing characteristics jointly complicate stable extraction of balance-relevant pose dynamics from raw temperature maps.

\noindent\textbf{\ding{183} Body balance quantification.}
Body balance is commonly categorized into three stages, \ie, stable balance (SB), loss of balance (LoB), and ground impact stage (GIS) \cite{yu2021large, zhang2020human}, as illustrated in \fig\ref{fig:balance_fall_overview}(third row).
Among these, the LoB phase is transient and subtle, making precise annotation difficult even with synchronized camera recordings.
Furthermore, pose estimates derived from thermal data are sensitive to subject direction, distance, and environmental layout, introducing uncertainty into body balance state classification.
These factors collectively challenge both ground-truth labeling and model-based inference of balance states.

\noindent\textbf{\ding{184} False-alarm suppression.}
In real-world deployments, fall detectors inevitably encounter out-of-vocabulary (OOV) behaviors arising from diverse daily activities and inter-subject variability.
Many non-fall movements may resemble imbalance patterns, leading to frequent false alarms, particularly for unseen activities, users, and environments. 
Suppressing such OOV-induced false alarms is therefore essential for reliable long-term operation.

\begin{figure}[t]
    \centering
    \includegraphics[width=0.99\linewidth]{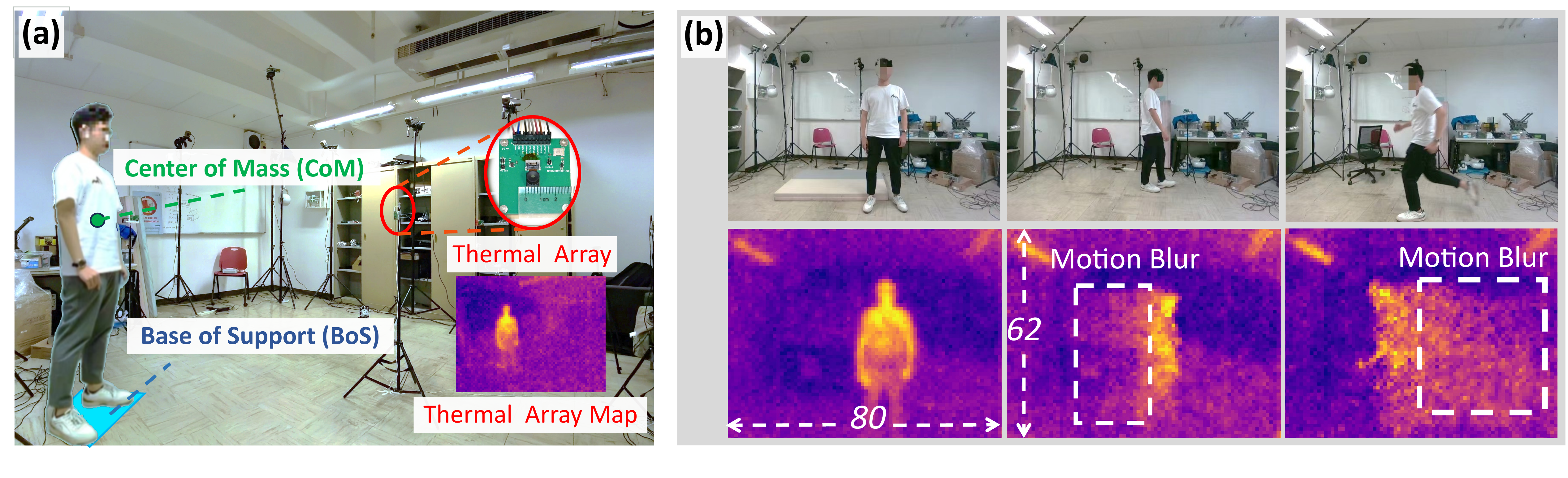}
    \caption{Thermal array sensing characteristics and biomechanical context. 
    \rm{(a) Operating principle of a thermal array sensor, which captures thermal radiation emitted by the human body to generate a low-resolution temperature map. The standing subject is annotated with the Center of Mass (CoM) and Base of Support (BoS). 
    (b) Comparison of static and dynamic postures under varying motion velocities. Thermal arrays produce stable temperature maps under static conditions, whereas rapid motion induces pronounced thermal motion blur (highlighted by white boxes), a phenomenon largely absent in RGB imagery. The progressive expansion of the highlighted regions indicates that motion blur magnitude increases with movement velocity.}}
   
    \label{fig:sensing_principle}
\end{figure}

To address these challenges, \sysname introduces three key novel techniques:

\head{$\blacksquare$ Appearance–motion fusion for pose estimation}
We exploit the observation that the human body exhibits relatively stable temperature distributions, with hotter regions at the head and exposed limbs, cooler regions around the clothed torso, and lower temperatures near the lower body.
These temperature appearances provide coarse yet reliable spatial cues even under low resolution as illustrated in \fig\ref{fig:sensing_principle}(a).
In addition, rather than treating motion blur (\fig\ref{fig:sensing_principle}(b)) as a nuisance, we reinterpret it as a physically meaningful motion cue: rapid movements smear thermal energy across neighboring pixels, encoding the magnitude and direction of velocity.
By fusing instantaneous temperature appearance cues with velocity- and direction-aware motion representations, \sysname achieves robust pose reconstruction even under rapid motion, self-occlusion, and severe resolution constraints.

\head{$\blacksquare$ Physically grounded balance-aware learning}
Discrete balance-state annotations (SB, LoB, GIS) are inherently ambiguous, particularly around the transient LoB phase.
To address this limitation, we introduce a continuous and physically grounded balance representation, termed the \textit{Signed Margin of Balance (SMoB)}, inspired by Winter’s biomechanical model of postural control \cite{winter2009biomechanics}. 
The ground truth SMoB is computed from 3D poses obtained from a separate motion-capture system.
By jointly leveraging continuous balance representation and discrete fall labels as supervision signals in a multi-task framework, \sysname captures subtle imbalance evolution more faithfully.

\head{$\blacksquare$ Pose-bridged OOV enhancement}
Beyond balance estimation, pose sequences provide an additional benefit as a bridge between thermal–based fall detection and large-scale public human activity datasets containing 3D pose annotations.
To enhance robustness against OOV behaviors and inter-subject variability, we pretrain \sysname using a large-scale human pose dataset, OctoNet \cite{yuanoctonet}, alongside our thermal array fall dataset.
Leveraging the 3D pose representations, we further introduce viewpoint-based data augmentation via randomized virtual camera projections and adopt a contrastive learning objective to promote viewpoint-variant robustness.
This pose-bridged OOV enhancement substantially improves generalization across unseen activities, users, and environments, as validated experimentally.

We prototype \sysname using a commercial thermal array sensor (Meridian MI0802M6S) that provides an $80\times62$ temperature map with a $90^{\circ}\times67^{\circ}$ FoV.
Each sensor node integrates the thermal array sensor with an ESP32 microcontroller for sensor polling, resulting in a total hardware cost of 18.35~USD.
Thermal maps are streamed from the ESP32 to a laptop equipped with an NVIDIA RTX 4080 GPU and an Intel i7-13650 CPU, where \sysname performs inference in real time at 20~Hz.
To evaluate \sysname, we collected a large-scale dataset comprising over 3,000 fall instances from 35 participants across three indoor environments: a laboratory, a meeting room, and a clinical simulation room.
\sysname achieves a detection rate (DR) of 98.26\% and a false alarm rate (FAR) of 0.65\%.
To further assess long-term robustness, we deployed \sysname continuously across four real-world indoor environments, including three bedrooms and one living room, for 27 days. 
The field trial demonstrated an ultra-low false alarm rate of 0.00126\%, confirming its suitability for unattended, long-term use. 
Considering the bathroom being the most dangerous place in the house for fall accidents, we further conduct a pilot study in a typical bathroom environment, successfully detecting all 25 falls with only a single false alarm, highlighting robustness under challenging moisture and temperature conditions.

\noindent\textbf{Contributions.}
\begin{itemize}[noitemsep, topsep=0pt, leftmargin=*]
\item We propose \sysname, to our knowledge, the first thermal-based fall detection system grounded in human balance dynamics, moving beyond the coarse motion cues that dominate prior approaches.
\item We introduce three key technical components: (i) appearance-motion fusion for robust pose reconstruction from low-resolution, motion-blurred thermal data; (ii) a physically grounded balance-aware learning framework based on the Signed Margin of Balance; and (iii) a pose-bridged OOV enhancement strategy that leverages large-scale public human pose datasets to improve generalization and suppress OOV-induced false alarms.
\item We validate \sysname through large-scale experiments and 27-day real-world deployments across four indoor environments, achieving high detection performance (98.26\% DR and 0.65\% FAR) and an ultra-low false alarm rate of 0.00126\% in long-term deployment, including challenging bathroom scenarios characterized by moisture and thermal interference.
\end{itemize}

\section{Primer}
\label{sec:primer}

In this section, we first outline the operating principles of thermal array sensors.
We then introduce the biomechanical foundations of human balance formulation.
Finally, we describe the motion blur artifacts inherent to the thermal array sensing process.

\subsection{Thermal Array Sensor}
As illustrated in \fig\ref{fig:sensing_principle}(a), a thermal array is a compact sensor composed of a grid of thermopile detectors that measure the thermal radiative power emitted by objects within its FoV.
The sensor produces a low-resolution temperature map (typically below $100 \times 100$), which is sufficient to capture coarse human pose dynamics while inherently preserving privacy by avoiding the appearance-level details present in RGB imagery.
Owing to their thermopile-based design, thermal arrays are substantially more cost-effective than high-resolution thermal cameras based on microbolometer or photon-detector technologies.
For example, the Meridian MI0802M6S thermal array sensor used in \sysname has a compact form factor of $9\,\mathrm{mm}\times 9\,\mathrm{mm}\times 8\,\mathrm{mm}$ and provides an $80 \times 62$ temperature map at a cost of 10~USD, whereas even entry-level thermal imaging modules, such as the Seek Thermal Mosaic Core, with a resolution of $320 \times 240$, typically cost around 500~USD.

This favorable trade-off among spatial resolution, cost, and intrinsic privacy preservation, together with reliable operation under no-light conditions, makes thermal array sensors an attractive modality for ubiquitous human sensing applications \cite{zhang2024tadar}.

\subsection{Biomechanical Foundations of Human Balance}
Human balance is fundamentally governed by the mechanical relationship between the body’s mass distribution and its support conditions under gravity.
From a biomechanical perspective, the human body balance is determined by the spatial relationship between the center of mass (CoM) and the base of support (BoS), as shown in \fig\ref{fig:sensing_principle}(a).

According to Winter’s theory \cite{winter2009biomechanics}, postural balance is maintained when the projection of the CoM lies within the BoS, whereas LoB emerges as the CoM projection approaches or exceeds the BoS boundary.
The CoM can be computed by modeling the body as a collection of anatomical segments.
For a segment $s$, its local center of mass $\mathbf{c}_s$ is defined as
$\mathbf{c}_s = \sum_{k=1}^{K_s} \alpha_{s,k}\,\mathbf{p}_{s,k}$,
where $K_s$ denotes the number of joints in the segment, $\mathbf{p}_{s,k}$ denotes the 3D position of the $k$-th joint, and $\alpha_{s,k}$ represents the segment-specific geometric weighting coefficient, satisfying $\sum_{k=1}^{K_s}\alpha_{s,k}=1$.
The whole-body CoM is then obtained as a mass-weighted aggregation of segment centers:
$\mathbf{CoM} = \sum_{s=1}^{S} m_s\,\mathbf{c}_s$,
where $S$ denotes the total number of body segments and $m_s$ denotes the mass fraction of segment $s$ based on standard anthropometric data \cite{winter2009biomechanics}.
The BoS is defined as a support polygon formed by the planar projection of ground-contact joints, typically approximated by the convex hull of foot contact points, as shown in \fig\ref{fig:sensing_principle}(a).
Together, the CoM and BoS are determined by the underlying 3D body pose and provide a biomechanical basis for characterizing human balance.
While the CoM–BoS relationship offers a qualitative description of balance states, we extend it to a continuous balance representation in \S\ref{subsec:balance_learning} to enable precise quantitative analysis.

\subsection{Motion Blur}
Motion blur in thermal arrays arises from the relatively slow thermal–electrical response of thermopile pixels, causing the observed frame $g$ to be a velocity-dependent convolution of an underlying sharp frame $f$.

As shown in \fig\ref{fig:sensing_principle}(b), temperature maps captured by the MI08 thermal array appear relatively sharp under static postures, whereas dynamic movements introduce pronounced motion blur.
The blur formation process in thermal arrays can be modeled as $g = f * h_v + n$, where $n$ denotes thermal noise and $h_v$ is a velocity-dependent point spread function (PSF) \cite{oswald2017motion}.
Specifically, the PSF is given by $h_v(x) = \frac{1}{v r \tau_{\mathrm{camera}}} \exp\!\left(-\frac{x}{v r \tau_{\mathrm{camera}}}\right)$, where $v$ denotes the motion velocity, $r$ the spatial resolution in meters per pixel, and $\tau_{\mathrm{camera}}$ the sensor response time.
This formulation indicates that stronger motion blur corresponds to higher movement velocity, implying that blur patterns encode meaningful temporal dynamics rather than acting solely as image degradation artifacts.
As illustrated in \fig\ref{fig:sensing_principle}(b), stationary subjects exhibit negligible blur, whereas moving subjects induce blur whose magnitude scales proportionally with motion speed.
Consequently, high-velocity activities such as running generate substantially more pronounced blur patterns than lower-velocity actions such as walking.

\begin{figure*}[t]
    \centering
    \includegraphics[width=0.99\textwidth]{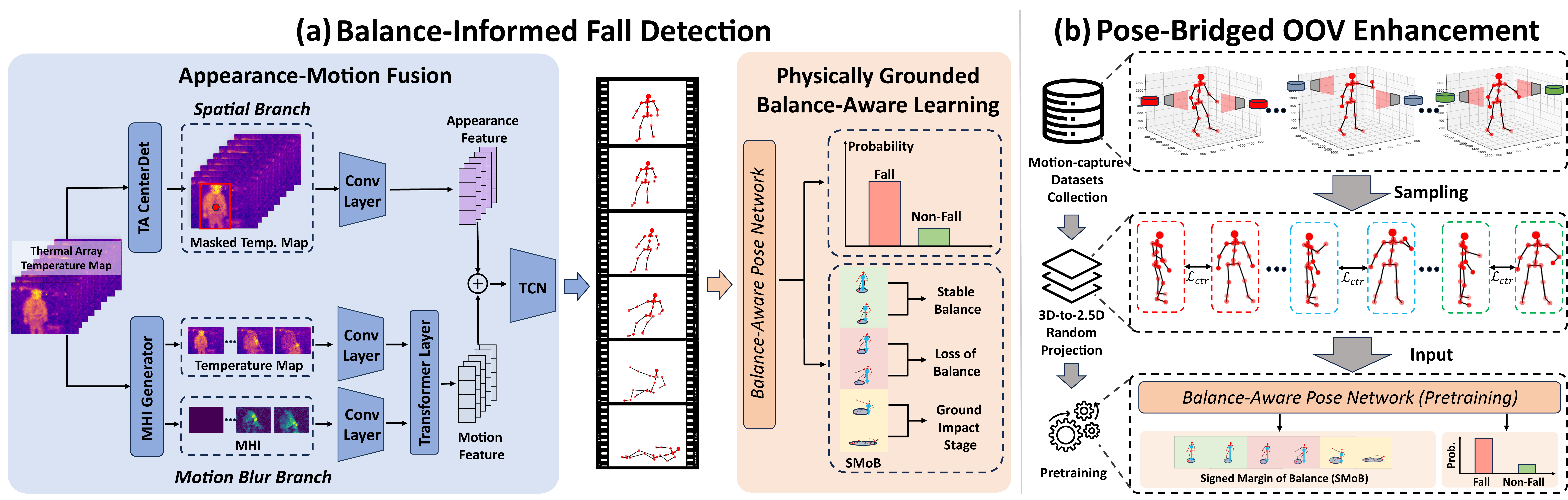}
    \caption{Overview of the \sysname framework. 
    \sysname consists of two complementary phases. 
    (a) \textbf{Balance-Informed Fall Detection}: an \emph{Appearance--Motion Fusion} module estimates a robust 2.5D human pose sequence from thermal array temperature maps by jointly leveraging spatial appearance cues and motion-blur–induced dynamics. 
    The resulting pose sequence is then processed by a \emph{Balance-Aware Pose Network}, which learns a \textit{Physically  grounded balance representation} to jointly infer continuous balance states and detect fall events. 
    (b) \textbf{Pose-Bridged OOV Enhancement}: the balance-aware pose network is pretrained using large-scale public pose data projected into diverse 2.5D views, with cross-view contrastive learning enforcing motion-sensitive and viewpoint-invariant representations, thereby enhancing robustness and reducing false alarms in real-world deployments.}
    
    \label{fig:sys_overview}
\end{figure*}

\section{\sysname Design}
\label{sec:design}

This section presents the overall architecture of \sysname and details its key components, spanning model design and practical implementation.

\subsection{Overview}
\sysname is designed to detect fall events from thermal array temperature maps, as illustrated in \fig\ref{fig:sys_overview}.
Its architecture comprises two complementary phases: (i) balance-informed fall detection (\fig\ref{fig:sys_overview}(a)) and (ii) pose-bridged OOV enhancement (\fig\ref{fig:sys_overview}(b)).

In the balance-informed fall detection phase, \sysname is trained on our self-collected fall detection dataset.
It is formulated as an end-to-end framework consisting of two core modules: an appearance-motion fusion module and a physically grounded balance-aware learning module.
The appearance-motion fusion module adopts a dual-path architecture to infer human pose sequences from thermal array temperature maps.
The physically grounded balance-aware learning module is implemented as a multi-task component that trains the balance-aware pose network to jointly model balance dynamics and fall events based on the inferred pose sequences.

In the pose-bridged OOV enhancement phase, we introduce a pose-bridged pretraining strategy to improve the robustness of \sysname against OOV behaviors, inter-subject variability, and viewpoint changes.
In this phase, the balance-aware pose network is pretrained using large-scale public pose data with cross-view contrastive learning enforcing motion-sensitive and viewpoint-invariant representations to enhance generalization.

\begin{figure}[t]
    \centering
    \includegraphics[width=0.8\linewidth]{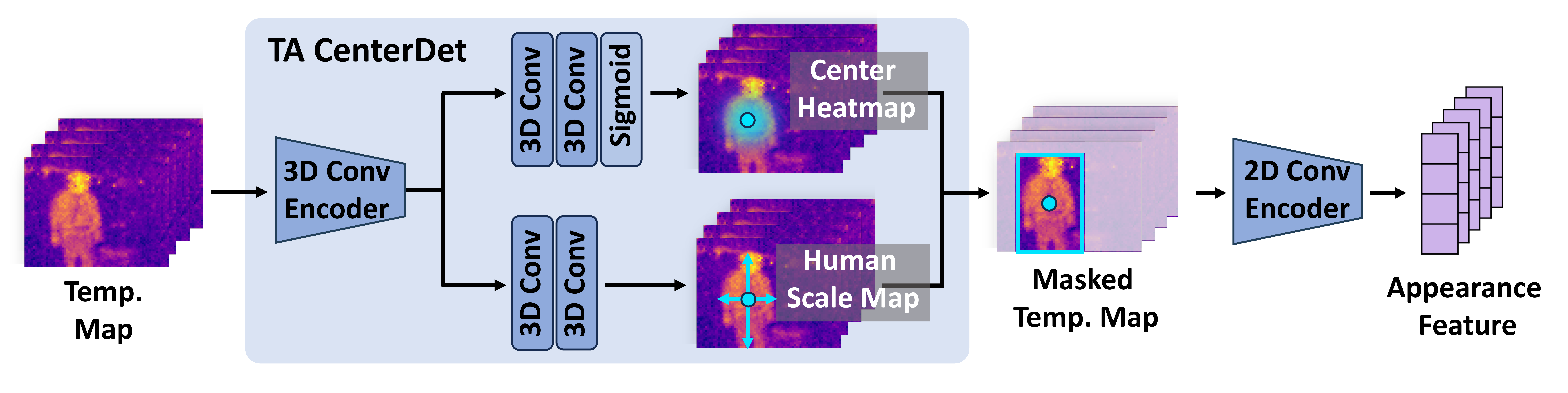}
    \caption{The structure of the Spatial Branch. It first predicts the Center Heatmap and the Human Scale Map from the thermal sequence. The two outputs are then combined to generate a Masked Temp Map, which is further processed by convolution layers to extract the \textit{Appearance Feature}.}
    \label{fig:appearance_branch}
\end{figure}

\subsection{Appearance-Motion Fusion}
Accurate pose estimation is a prerequisite for reliable inference of human balance states, as established in \S\ref{sec:primer}.
In static scenarios, the human body exhibits relatively stable thermal distributions: warmer regions typically correspond to the head and exposed extremities, the torso appears cooler due to clothing insulation, and the lower body often presents lower temperatures.
Exploiting these consistent spatial patterns, we introduce a \textit{Spatial Branch} to extract discriminative \textit{appearance features} from thermal observations.
In dynamic scenarios, rapid limb or torso movements induce thermal motion blur, which encodes physically meaningful information about underlying motion, including velocity magnitude and movement direction, as discussed in \S\ref{sec:primer}.
Accordingly, we introduce a dedicated \textit{Motion Blur Branch} to extract complementary motion features.
Finally, appearance and motion features from the two branches are fused to estimate human pose, enabling robust pose inference across both static and dynamic conditions.

\subsubsection{Spatial branch}
The spatial branch focuses on extracting appearance-based cues from thermal observations, which are particularly reliable in static or slowly varying scenarios.

To extract the temperature appearances, we first process the thermal frames with \textit{TA-CenterDet}, a lightweight center-based detector inspired by CenterNet~\cite{duan2019centernet}.
As illustrated in \fig\ref{fig:appearance_branch}, the input \textit{Temp. Map} sequence is first encoded by a 3D Convolutional Encoder to capture spatiotemporal thermal features.
These features are then processed by a series of 3D Convolutional Layers, followed by a sigmoid operation, to generate two complementary outputs: the \textit{Center Heatmap} and the \textit{Human Scale Map}.
The \textit{Center Heatmap} indicates the most probable human center in each frame, while the \textit{Human Scale Map} characterizes the spatial extent of the human body around the predicted center.
By jointly leveraging the \textit{Center Heatmap} and the \textit{Human Scale Map}, TA-CenterDet constructs a \textit{Masked Temp. Map} that localizes the human-centered region and suppresses background thermal interference.
By linking the frame-wise localization results over time, \textit{TA-CenterDet} further forms a spatio-temporal bounding-box tube that consistently tracks the human region across the thermal sequence.

Finally, the resulting \textit{Masked Temp. Map} is fed into a 2D Convolutional Encoder to extract the \textit{Appearance Feature}.
This representation, denoted as $\mathbf{f}^{\mathrm{ap}}_t$, captures structural cues such as coarse body contours and stable thermal gradients.
Since these cues depend on intact spatial patterns, the appearance representation may degrade under fast motion, where thermal measurements exhibit strong blur.

\begin{figure}[t]
    \centering
    \includegraphics[width=0.99\linewidth]{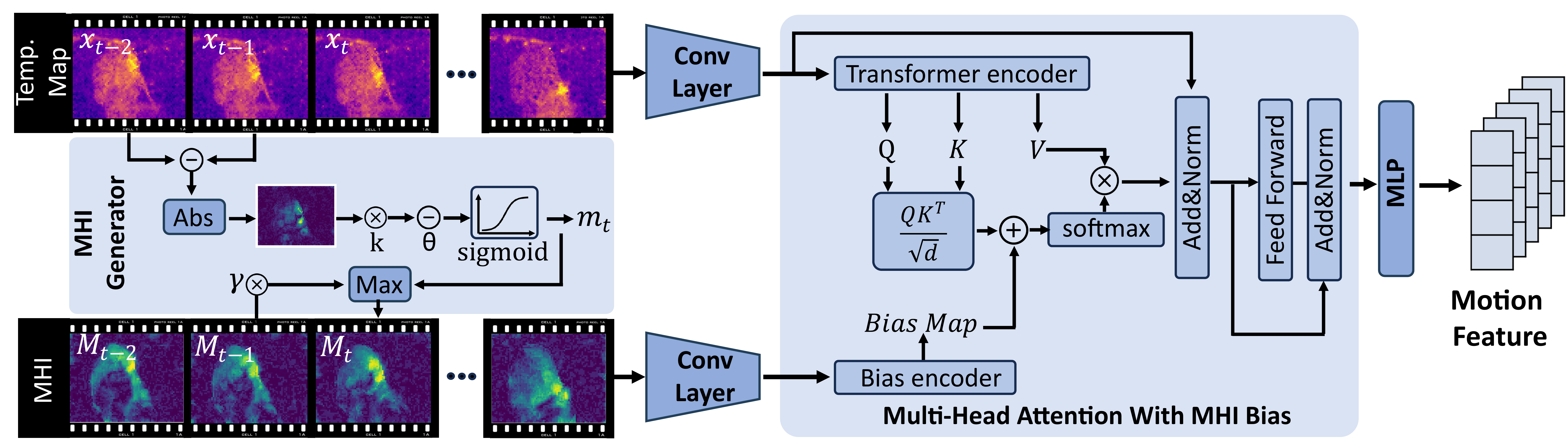}
    \caption{The structure of Motion Blur Branch. It first generates the \textit{Motion History Image} and then uses an attention mechanism to extract \textit{motion feature} from temperature map and motion history image.}
    \label{fig:appearance-motion_fusion}
\end{figure}

\subsubsection{Motion blur branch}
Building upon the observation that thermal motion blur encodes informative motion dynamics, this branch aims to explicitly extract and leverage such cues for pose estimation under rapid movements.
Specifically, this branch contains two key components: Motion History Image (MHI) Generator and MHI Biased Multi-head Attention.

\ding{182} MHI Generator. To explicitly extract motion dynamics, we compute MHI sequence \cite{bobick2002recognition} from the original thermal maps.
As shown in \fig\ref{fig:appearance-motion_fusion}, the MHI provides a compact temporal template that emphasizes regions undergoing continuous motion, making it particularly suitable for thermal array data, where motion blur naturally spreads thermal energy along the direction of movement.
By integrating frame-to-frame variations into a single representation, the MHI selectively amplifies pixels affected by recent motion while attenuating static or slowly varying regions, thereby isolating the spatial footprints created by high-velocity body parts.
Let $x_t$ denote the normalized temperature map at time~$t$.
We first compute a \textit{soft motion mask} based on inter-frame differences:
\begin{equation}
m_t = \sigma\!\left(k\bigl(|x_t - x_{t-1}| - \theta\bigr)\right),
\end{equation}
where $\sigma(\cdot)$ is the sigmoid function, $k$ controls the sharpness of the soft threshold, and $\theta$ determines the motion activation level.
Motion information is then accumulated over time using an exponential decay scheme:
\begin{equation}
M_t = \max\!\bigl(\gamma M_{t-1},\; m_{t-1}\bigr),
\end{equation}
where $M_t$ denotes the MHI at time~$t$ and $\gamma \in (0,1)$ controls the fading rate of past motion responses.
This formulation highlights recent motion while naturally suppressing older activity, yielding a temporally smoothed representation of movement dynamics.

\ding{183} MHI Biased Multi-head Attention. To incorporate motion cues into pose estimation, as shown in \fig\ref{fig:appearance-motion_fusion}, the MHI is processed by a lightweight convolutional network to generate an attention bias map $b_{\mathrm{MHI}}$.
In parallel, the original temperature maps are encoded by a convolutional backbone and projected to form the query, key, and value representations $(q,k,v)$ for a spatial Transformer.
The motion-induced bias is injected into the attention computation as
\begin{equation}
\mathrm{Attn}(q,k)
  = \mathrm{softmax}\!\left(\frac{qk^{\top}}{\sqrt{d_h}} + b_{\mathrm{MHI}}\right),
\end{equation}
where $d_h$ is the dimensionality of each attention head.
By introducing $b_{\mathrm{MHI}}$ as an additive bias, the transformer encoder is explicitly guided to attend more strongly to velocity-dominant regions, while suppressing spatial locations with weak or ambiguous motion cues.
The attention-weighted output yields the \textit{motion feature} $\mathbf{f}^{\mathrm{mhi}}_t$, which remains robust in dynamic scenarios where appearance features derived from raw temperature maps are degraded by motion blur.

\subsubsection{Pose regression}
The representations from both branches are fused through element-wise addition to obtain a unified descriptor: $\mathbf{f}_t = \mathbf{f}^{\mathrm{ap}}_t + \mathbf{f}^{\mathrm{mhi}}_t.$
The sequence $\{\mathbf{f}_t\}_{t=1}^T$ is refined by a Temporal Convolutional Network (TCN) \cite{lea2016temporal}, which enforces temporal continuity and stabilizes joint trajectories.  
Finally, a linear regression head maps the refined temporal features to a 2.5D pose sequence 
$\mathbf{S} \in \mathbb{R}^{B \times T \times J \times 3}$,  
where each joint contains its $(x,y)$ image-plane coordinates and a depth value.

\begin{figure}[t]
    \centering
    \includegraphics[width=0.99\linewidth]{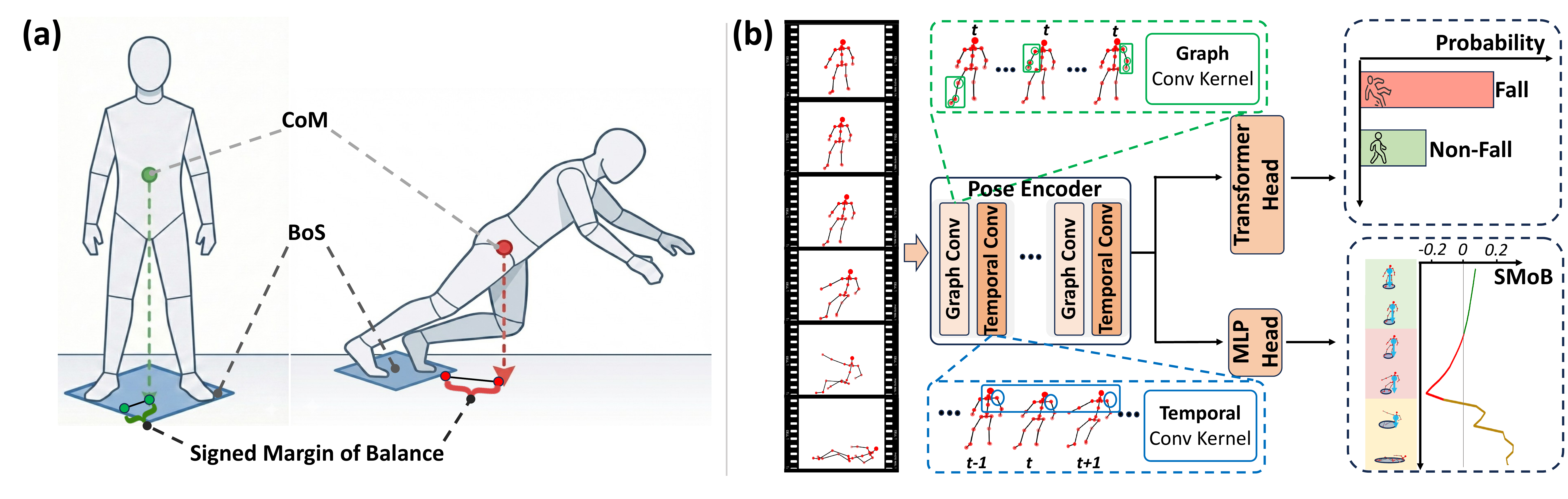}
    \caption{The illustration of the Physically Grounded Balance Representation and the Balance-Aware Pose Network. (a) \textbf{Physically Grounded Balance Representation}: Balance is quantified using the Signed Margin of Balance (SMoB), defined as the signed distance between the projected center of mass (CoM) and the boundary of the base of support (BoS). (b) \textbf{Balance-Aware Pose Network}: The predicted pose sequence is encoded using ST-GCN-based pose encoder to capture postural evolution. The encoded pose representation is then fed into two lightweight prediction branches: a temporal transformer head for fall-event classification, and an MLP head for SMoB prediction. }
    \label{fig:balance_representation}
\end{figure}

\subsection{Physically Grounded Balance-Aware Learning}
\label{subsec:balance_learning}
In this subsection, we detail how the physically grounded balance representation is defined and the training process of the Balance-Aware Pose Network.

\subsubsection{Physically grounded balance representation}
After estimating human pose from thermal observations, the next critical step is to extract features that are directly related to body balance dynamics.

Generally, the body balance dynamics can be described by three canonical balance states, \ie, Stable Balance (SB), Loss of Balance (LoB), and Ground Impact Stage (GIS).
Following this, a fall is modeled as a temporal progression through these states, typically following the sequence SB $\rightarrow$ LoB $\rightarrow$ GIS.
Accordingly, a balance-state sequence can be represented as $\{SB_{0}\ldots SB_{i},\; LoB_{0}\ldots LoB_{j},\; GIS_{0}\ldots GIS_{k}\}$
where the $i$, $j$ and $k$ represent the number of the SB, LoB and GIS frame respectively. 
While GIS frames can be reliably annotated through visual inspection, the boundary between SB and LoB is inherently ambiguous.
Existing works often define the onset of a fall based on acceleration peaks~\cite{yu2021large}, which primarily capture motion intensity rather than mechanical balance.
Such definitions fail to explicitly reflect whether the body remains dynamically stable.

To address this limitation, as shown in \fig\ref{fig:balance_representation}, we propose a Physically grounded balance representation, named \textit{Signed Margin of Balance (SMoB)} following Winter’s biomechanical formulation as introduced in \S\ref{sec:primer}.
SMoB is derived from the classical Margin of Stability (MoS) \cite{hof2005condition} and serves as a simplified, balance-centric representation tailored for fall detection, capturing body balance through the spatial relationship between the projected CoM and the BoS.
Specifically, when the CoM projection remains within the BoS, the body can generate sufficient restoring torque to maintain equilibrium;  
conversely, when the CoM moves close to or beyond the BoS boundary, denoted as $\partial BoS$, the restoring moment becomes insufficient, indicating a loss of balance.  
Quantitatively, the SMoB is computed as a signed postural balance margin, which measures the shortest horizontal distance between the projected CoM and the boundary of the BoS:  
\begin{equation}
d = \min_{p \in \partial BoS}\|\mathrm{proj(CoM)} - p\|,
\label{eq:psi_distance}
\end{equation}

\begin{equation}
\mathrm{SMoB} =
\begin{cases}
+d, & \text{if } \mathrm{proj(CoM)} \in BoS,\\[4pt]
-d, & \text{otherwise.}
\end{cases}
\label{eq:psi_signed}
\end{equation}
A positive $\mathrm{SMoB}$ indicates that the CoM projection remains within the BoS and the posture is mechanically stable, 
whereas a negative $\mathrm{SMoB}$ signifies that the CoM has moved beyond the support boundary, marking the onset of imbalance.  
This signed formulation provides a continuous and physically interpretable measure of postural balance throughout the motion sequence.

Unlike discrete balance-state annotations, the continuous SMoB trajectory captures both gradual imbalance accumulation and abrupt transitions, providing a fine-grained characterization of balance dynamics during pre-fall and impact (after-fall) phases.
However, SMoB requires accurate 3D pose information for the BoS and CoM calculation (refer to \S\ref{sec:primer}), which is difficult to obtain reliably from low-resolution thermal observations alone.
Hence, we compute the SMoB using motion capture–based ground-truth poses, and employ it as an auxiliary supervision signal within a multitask learning framework to train the Balance-Aware Pose Network as detailed below.

\subsubsection{Balance-aware pose network training}
The Balance-Aware Pose Network is designed to learn a compact and balance-sensitive representation from pose dynamics for robust fall detection.
As shown in \fig\ref{fig:balance_representation}(b), it takes the 2.5D pose sequences as input to infer the SMoB and the fall event.
We will first detail the model structure, then, focus on the training strategy.

\ding{182} Model structure: 
As shown in \fig\ref{fig:balance_representation}(b), the Balance-aware pose network contains three components, a pose encoder, a transformer head for fall event detection, and a MLP head for SMoB estimation.
First, accurately characterizing human balance requires modeling not only instantaneous body posture but also the temporal evolution of joint configurations as imbalance accumulates.
Falls are typically preceded by coordinated biomechanical patterns, such as progressive forward lean, narrowing of the base of support, or failed compensatory arm motions.
These phenomena arise from strong kinematic coupling among body segments and cannot be captured by treating joints or frames independently.
To this end, we encode pose dynamics using the pose encoder, as a Spatiotemporal Graph Convolutional module (ST-GCN)~\cite{yan2018spatial}, which represents each pose frame as a graph with joints as nodes and bones as edges.
By jointly modeling spatial joint dependencies and temporal motion patterns, the ST-GCN produces a latent pose representation
$\mathbf{H}_{\mathrm{pose}} = g_{\theta}(\mathbf{S}) \in \mathbb{R}^{B \times T \times 256},$
where $\mathbf{S} \in \mathbb{R}^{B \times T \times J \times 3}$ represents the input pose sequence,  $g_{\theta}$ denotes the ST-GCN pose encoder.
This representation emphasizes postural evolution over time and is particularly sensitive to subtle configuration changes preceding the loss of balance.

Then, to capture long-term temporal dependencies for accurate fall detection beyond the local receptive field of ST-GCN, 
we apply a lightweight temporal Transformer head on top of $\mathbf{H}_{\mathrm{pose}}$.
The Transformer head aggregates pose dynamics over the full sequence via self-attention and outputs the fall probability
$\hat{y} = f_{\mathrm{cls}}(\mathbf{H}_{\mathrm{pose}})$.
In parallel, to enforce balance-aware representation learning, we attach a MLP head to estimate the SMoB at each time step:
$\hat{\mathrm{SMoB}}_{t} = f_{\mathrm{bal}}(\mathbf{H}_{\mathrm{pose},t}).$
By jointly regressing SMoB, the encoder is encouraged to focus on biomechanically meaningful pose dynamics rather than appearance-driven motion alone.

\ding{183} Training strategy: the Balance-aware pose network is trained with multi-task objective combines fall classification and balance regression:
\begin{equation}
\mathcal{L}
=
\mathcal{L}_{\mathrm{cls}}
+
\lambda \, \mathcal{L}_{\mathrm{bal}},
\label{eq:multi_task_loss}
\end{equation}
where $\mathcal{L}_{\mathrm{cls}}$ is the cross-entropy loss for fall/non-fall classification,
$\mathcal{L}_{\mathrm{bal}}$ is the mean squared error loss for SMoB prediction detailed below,
and $\lambda$ controls the contribution of balance supervision.

Formally, given the estimated balance representation sequence 
$\{\mathrm{SMoB}_t\}_{t=1}^{T}$,  
we define a balance–regression target $y^{\mathrm{bal}}_t = \mathrm{SMoB}_t,$
and supervise the model with a mean-squared error loss:
\begin{equation}
\mathcal{L}_{\mathrm{bal}}
= \frac{1}{T}\sum_{t=1}^{T}
\bigl(\hat{y}^{\mathrm{bal}}_t - y^{\mathrm{bal}}_t \bigr)^2,
\label{eq:balance_mse}
\end{equation}
where $\hat{y}^{\mathrm{bal}}_t$ denotes the predicted SMoB.  
Notably, we do not require explicit SB/LoB classification for training; instead, we directly regress the signed SMoB values as an auxiliary target.
This physics-guided auxiliary regression task provides dense supervision 
for the temporal evolution of balance.

\begin{figure}[t]
    \centering
    \includegraphics[width=0.99\linewidth]{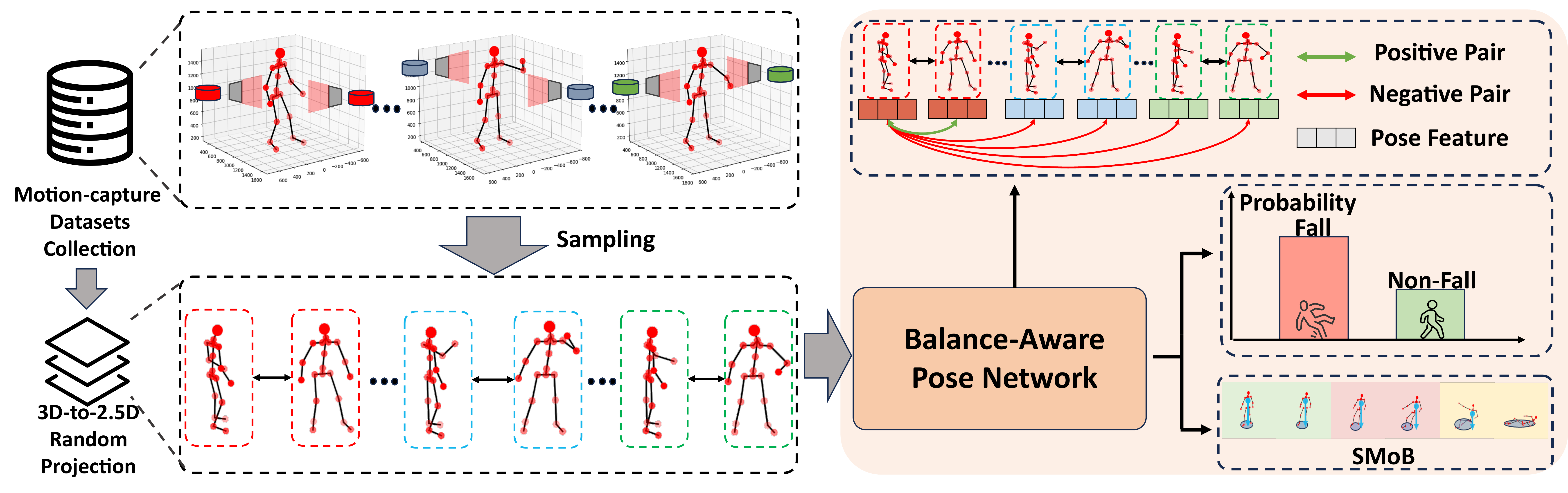}
    \caption{Illustration of the Pose-Bridged OOV Enhancement strategy. A motion-capture dataset provides 3D skeletal trajectories and is converted into the 2.5D coordinate space via \textit{3D-to-2.5D Random Projection} by sampling virtual camera viewpoints, producing multiple 2.5D projected views of the same motion sequence. The projected pose sequences are used to pretrain the \textit{Balance-Aware Pose Network} with a multi-objective formulation that combines fall classification, balance regression, and an additional \textit{View-Consistent Contrastive Learning} objective to encourage cross-view feature consistency.}

    \label{fig:pose-bridged_pretain}
\end{figure}

By integrating continuous balance dynamics with categorical fall-state 
semantics, our multi–task learning framework captures both 
\emph{how much} balance is lost and \emph{when} critical transitions occur, 
leading to more accurate and robust fall understanding.

\subsection{ Pose-Bridged OOV Enhancement}

To equip the balance aware pose network with stronger pose priors and improved robustness to OOV behaviors and inter-subject variations, we propose a pose-bridged OOV enhancement strategy that leverages both the OctoNet \cite{yuanoctonet} motion-capture dataset and our thermal-array fall data.
OctoNet \cite{yuanoctonet} provides accurate 3D skeletal trajectories covering 61 diverse non-fall daily activities, exposing the encoder to a broad range of natural human motions that are absent from thermal fall datasets and essential for handling OOV movements.
By jointly training on motion-capture trajectories from OOV behavior dataset and our own fall data, the pose encoder in the Balance-aware pose network learns motion-sensitive pose features that transfer reliably to the downstream fall detection task.
Besides, robustness to viewpoint variation is crucial for reliable fall detection in real deployments, where sensors are often installed at different locations and observe subjects from diverse angles. Changes in viewing direction can significantly alter the observed pose trajectories, especially under low-resolution thermal sensing, and may lead to inconsistent representations of the same underlying motion pattern.

To this end, as shown in \fig\ref{fig:pose-bridged_pretain}, our training objective augments the standard classification loss $\mathcal{L}_{\mathrm{cls}}$ and balance regression loss $\mathcal{L}_{\mathrm{bal}}$ with an additional contrastive learning term. 
Specifically, pose sequences corresponding to the same action but captured from different camera directions are treated as positive pairs, while unrelated sequences serve as negatives. 
This contrastive objective encourages the pose encoder to learn view-invariant motion representations, thereby reducing viewpoint-induced feature variations and improving generalization across diverse deployment configurations.

\head{3D-to-2.5D random projection}
Since the pretraining data are provided in full 3D world coordinates, we convert each motion sequence into our 2.5D thermal-sensing coordinate space by randomly sampling virtual camera viewpoints for data augmentation. 
Let a 3D motion sequence be $\mathbf{S}^{3D}=\{\mathbf{p}_{t,j}\in\mathbb{R}^{3} \mid t=1\dots T,\; j=1\dots J\}$.  
For each sequence, we independently sample two virtual camera locations $\mathbf{c}^{(1)},\mathbf{c}^{(2)} \sim \mathcal{U}$ and obtain two 2.5D projected views:
\begin{equation}
\mathbf{S}^{(v)} = \Pi\!\left(\mathbf{S}^{3D}, \mathbf{c}^{(v)}\right), \qquad v\in\{1,2\}.
\label{eq:two_view_projection}
\end{equation}
where $\Pi(\cdot)$ applies the projection $\mathbf{p}^{2.5D}_{t,j}=\Pi(\mathbf{p}_{t,j},\mathbf{c})$ to all joints and frames.  
These two projections yield different 2.5D views of the \emph{same} underlying 3D motion, serving as viewpoint augmentation and improving robustness to camera placement.

\head{View-Consistent Contrastive Learning}
To enforce cross-view consistency, we optimize the encoder with a contrastive objective over the two projected views of the same underlying 3D motion.  
The two projected views are encoded by the same backbone $g_{\theta}$ to obtain their pose features:
\begin{equation}
z^{(1)} = g_{\theta}\!\left(\mathbf{S}^{(1)}\right), \qquad
z^{(2)} = g_{\theta}\!\left(\mathbf{S}^{(2)}\right),
\label{eq:view_embeddings}
\end{equation}
where $\mathbf{S}^{(1)}$ and $\mathbf{S}^{(2)}$ are two 2.5D projections derived from the same underlying 3D motion sequence $\mathbf{S}^{3D}$.
A standard InfoNCE loss encourages the encoder to pull $(z^{(1)},z^{(2)})$ together while pushing apart pose features from different 3D sequences:
\begin{equation}
\mathcal{L}_{\mathrm{ctr}}
=
- \log
\frac{
\exp\left(\mathrm{sim}(z^{(1)},z^{(2)})/\tau\right)
}{
\sum_{k=1}^{N} \exp\left(\mathrm{sim}(z^{(1)},z_k)/\tau\right)
},
\label{eq:infonce}
\end{equation}
where $\mathrm{sim}(\cdot,\cdot)$ denotes cosine similarity, $\tau$ is a temperature hyper-parameter, and $\{z_k\}_{k=1}^{N}$ denotes the set of candidate representations in the mini-batch, containing the paired view $z^{(2)}$ and representations from other 3D motion sequences. This cross-view consistency serves as an effective motion prior, enabling the encoder to learn stable temporal patterns even under severe self-occlusion or viewpoint changes.

\begin{figure*}[t]
    \centering
    \includegraphics[width=0.99\textwidth]{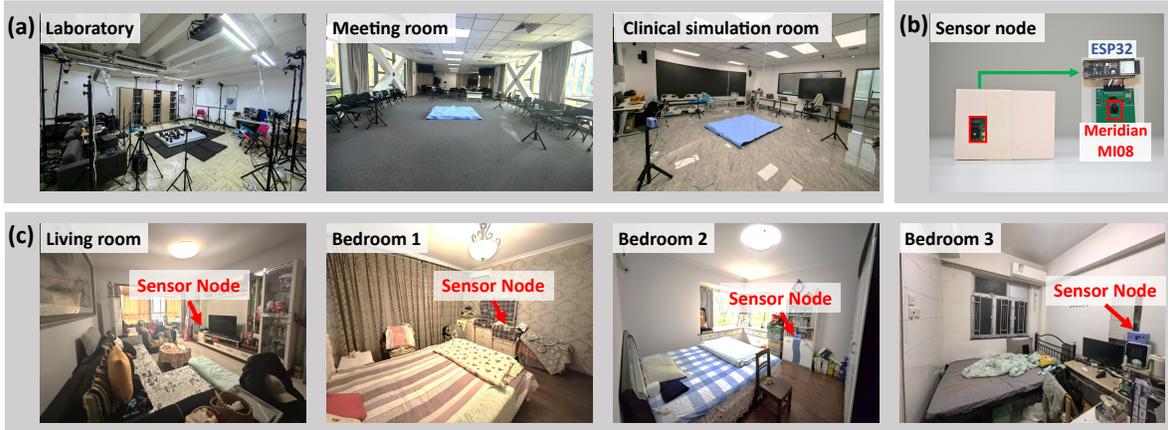}
    \caption{Experimental setup. \fig (a) illustrates the three representative indoor environments used for fall data collection. \fig (b) shows the Meridian MI0802M6S thermal array sensor deployed in all sessions. \fig (c) presents the four real-world scenarios for long-term evaluation. }
    \label{fig:experimental_setup}
\end{figure*}

\head{Pose encoder pretraining}
The pose encoder of \textit{Balance-Aware Pose Network} is pretrained with a multi-objective learning strategy that combines fall classification, balance regression, and cross-view contrastive supervision.

Given the projected pose sequence $\mathbf{S}$, the encoder predicts a fall label and an SMoB trajectory while also producing a latent representation for contrastive learning.  
The overall pretraining loss is
\begin{equation}
\mathcal{L}_{\mathrm{pre}}
=
\mathcal{L}_{\mathrm{cls}}
+
\lambda_{\mathrm{b}}\mathcal{L}_{\mathrm{bal}}
+
\lambda_{\mathrm{ctr}}\mathcal{L}_{\mathrm{ctr}},
\label{eq:pretrain_loss}
\end{equation}
where $\mathcal{L}_{\mathrm{cls}}$ is the binary fall classification loss,  
$\mathcal{L}_{\mathrm{bal}}$ is the mean squared error between predicted and ground-truth SMoB,  
and $\mathcal{L}_{\mathrm{ctr}}$ enforces cross-view feature consistency through contrastive learning.  
The weights $\lambda_{\mathrm{b}}$ and $\lambda_{\mathrm{ctr}}$ control the contributions of the balance supervision and the contrastive regularization.

After pretraining, the Balance-Aware Pose Network is initialized with the pretrained weights, and the intermediate layers of the pose encoder are frozen during fine-tuning to preserve the learned motion priors. This initialization equips the network with strong, physiology-aware motion priors and viewpoint-invariant representations, significantly improving robustness across fall directions, postural variations, and environment-induced sensing changes.

\section{Implementation}
\label{sec:impl}
\head{Hardware}
We implement \sysname using the Meridian MI0802M6S thermal array sensor, as illustrated in \fig\ref{fig:experimental_setup}(b).
The sensor outputs an $80\times62$ temperature map with a $90^{\circ}\times67^{\circ}$ FoV.
It is interfaced with an ESP32 microcontroller via the SPI bus, which performs sensor polling and frame packetization.
The ESP32 streams thermal frames over WiFi to a remote computing device for inference with \sysname.
In our prototype system, \sysname (51~GFLOPs per frame) runs on a laptop equipped with an NVIDIA RTX 4080 GPU and an Intel i7-13650 CPU, achieving real-time inference at 20~Hz.
The total hardware cost of a sensor node is approximately 18.35~USD, including 10~USD for the Meridian MI0802M6S and 8.35~USD for the ESP32.
Beyond the current implementation, \sysname can be further compressed through knowledge distillation and pruning, facilitating future edge deployment.
Moreover, \sysname maintains strong performance under reduced spatial resolution, as validated in our experiments, suggesting its applicability to other thermal array sensors with varying resolutions.
For scenarios requiring wider spatial coverage, the FoV can be extended to $105^{\circ}\times79^{\circ}$ using the MI0802M7G variant.

\head{Software}
The \sysname models are implemented in PyTorch \cite{paszke2019pytorch} and optimized using the AdamW optimizer.
Training is performed in a multi-stage manner to progressively stabilize learning across different modules.
The TA-CenterDet module is trained for human localization using a batch size of 16, a learning rate of 0.001, and AdamW optimization for 50 epochs.
Following this stage, the TA-CenterDet parameters are frozen, and the Appearance–Motion Fusion module is trained with a batch size of 16 and a learning rate of 0.0005 for 200 epochs.
With the Appearance–Motion Fusion module fixed, the balance-aware pose network is subsequently fine-tuned for fall detection using a batch size of 8 and a learning rate of 0.0005 for 40 epochs.
During this stage, balanced mini-batch sampling is adopted to address the class imbalance introduced by the sliding-window input strategy, ensuring an equal proportion of fall and non-fall samples during training.
To characterize model complexity, \sysname contains 11.84M learnable parameters.

\section{Experiments}
\label{sec:exp}

\subsection{Experimental Setup}
\head{Data Collection}
We collected fall data across three representative indoor environments: a laboratory, a meeting room, and a clinical simulation room, as shown in \fig \ref{fig:experimental_setup}(a). A total of 35 volunteers participated, resulting in 3,005 recorded fall instances. The participants were aged 20 to 39 years, with body weights of 35 to 80 kg and heights of 153 to 191 cm, including 12 males and 23 females.
In the lab environment, we gathered 2,764 fall instances from 15 volunteers, captured from 17 to 20 distinct sensor viewpoints. Participants in the lab sessions primarily wore short-sleeved clothing and were instrumented with motion-capture markers to enable accurate ground-truth recording. In the meeting room, we recorded 140 fall instances from 7 volunteers using 5 sensor nodes. In the clinical simulation room scenario, we collected 101 fall instances from 16 volunteers across 3 different views. 
In these two settings (meeting room and clinical simulation room), participants typically wore outerwear and did not wear motion-capture markers, better reflecting practical deployment conditions.
All volunteers were instructed to perform randomized fall sequences covering multiple fall types (\eg, slips, trips). Leveraging our multi-view thermal array setup, we captured falls from diverse orientations and trajectories. In addition to actual falls, we also collected fall-like activities—including sitting down, picking up objects, and lying on the bed—to improve robustness against false alarms. Across environments, the dominant background thermal interference mainly came from heat radiation of lighting fixtures and nearby electronic devices (\eg, computers), which introduces realistic nuisance thermal sources during data collection. All data collection protocols involving human participants were reviewed and approved by our university's Institutional Review Board (IRB), and informed consent was obtained from all participants prior to data collection.

\head{Train/Test Split}
We used 80\% of the data from the lab and meeting room for model training. The remaining 20\% of these data, together with all data collected in the clinical simulation room, were used exclusively for testing to evaluate cross-environment generalization.

\head{Metrics} To assess the performance of \sysname in fall detection, we evaluate the system using Detection Rate (DR) and False Alarm Rate (FAR) \cite{ji2022sifall}.
DR measures the proportion of true fall events that are correctly detected by the system.
FAR quantifies the proportion of non-fall activities that are incorrectly flagged as falls, reflecting the system's tendency to raise false alarms.

\head{Baselines}
To comprehensively evaluate the effectiveness of \sysname, we compare it with four representative baselines that span different modeling paradigms in thermal-array-based fall detection. 
Specifically, we include two existing methods tailored for thermal array sensing, which represent statistical and detection-driven approaches, as well as two sequence baselines that model fall dynamics from sequence data at different representation levels.

1) EMD \cite{newaz2025approach}: 
This baseline characterizes fall events by modeling the statistical distribution of thermal patterns captured by infrared array sensors.
Instead of learning discriminative representations through deep networks, it compares empirical thermal distributions with predefined statistical models using Earth Mover’s Distance (EMD).
Normal activities are identified by their alignment with Beta or Normal distributions, while fall events exhibit significantly larger distributional divergence.
This method represents distribution-level statistical modeling without relying on explicit pose or structural representations.

2) YOLO-CBAM \cite{jiang2024fall}: 
This baseline formulates fall detection as an object detection task using a modified YOLOv5 network to localize and classify fall-related patterns from infrared array maps.
It incorporates the Convolutional Block Attention Module (CBAM) to enhance spatial and channel-wise feature representation, together with lightweight architectural optimizations for efficient inference.
While effective at capturing mid-level structural cues from thermal images, it does not explicitly model temporal dynamics, pose structures, or biomechanical balance information.
It thus represents a detection-based approach driven by discriminative visual patterns rather than explicit human motion or balance modeling.

3) Seq-CNN:
This baseline directly classifies sequences of low-resolution infrared maps using convolutional neural networks. Each frame is processed by a CNN to extract spatial features, followed by temporal modeling using global average pooling across time. It serves as a representative of pixel-level modeling without relying on explicit structural representations. This design is consistent with existing thermal-array-based fall detection methods that directly use raw thermal maps as network inputs without explicit detection or pose estimation \cite{adolf2018deep, rezaei2021unobtrusive}.

\begin{figure*}[t]
\begin{minipage}[t]{0.285\textwidth}
  \vspace{0pt}
    \centering
      \includegraphics[width=0.99\textwidth]{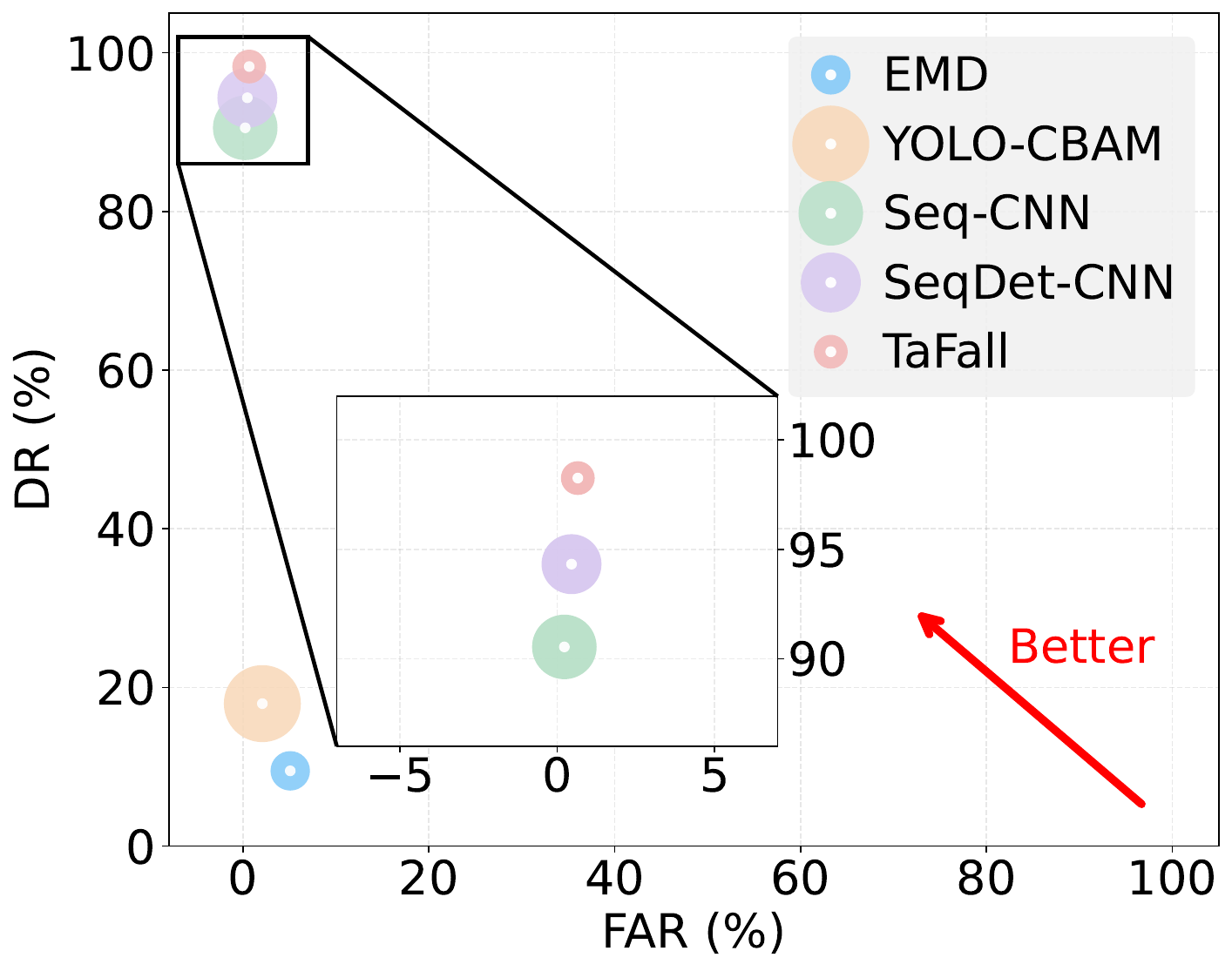}
    \vspace{-1.8\baselineskip}
    \caption{Overall performance.}
    \label{fig:overall_performance_distance_f1}
  \end{minipage}
\begin{minipage}[t]{0.27\textwidth}
  \vspace{0pt}
    \centering 
    \includegraphics[width=0.99\textwidth]{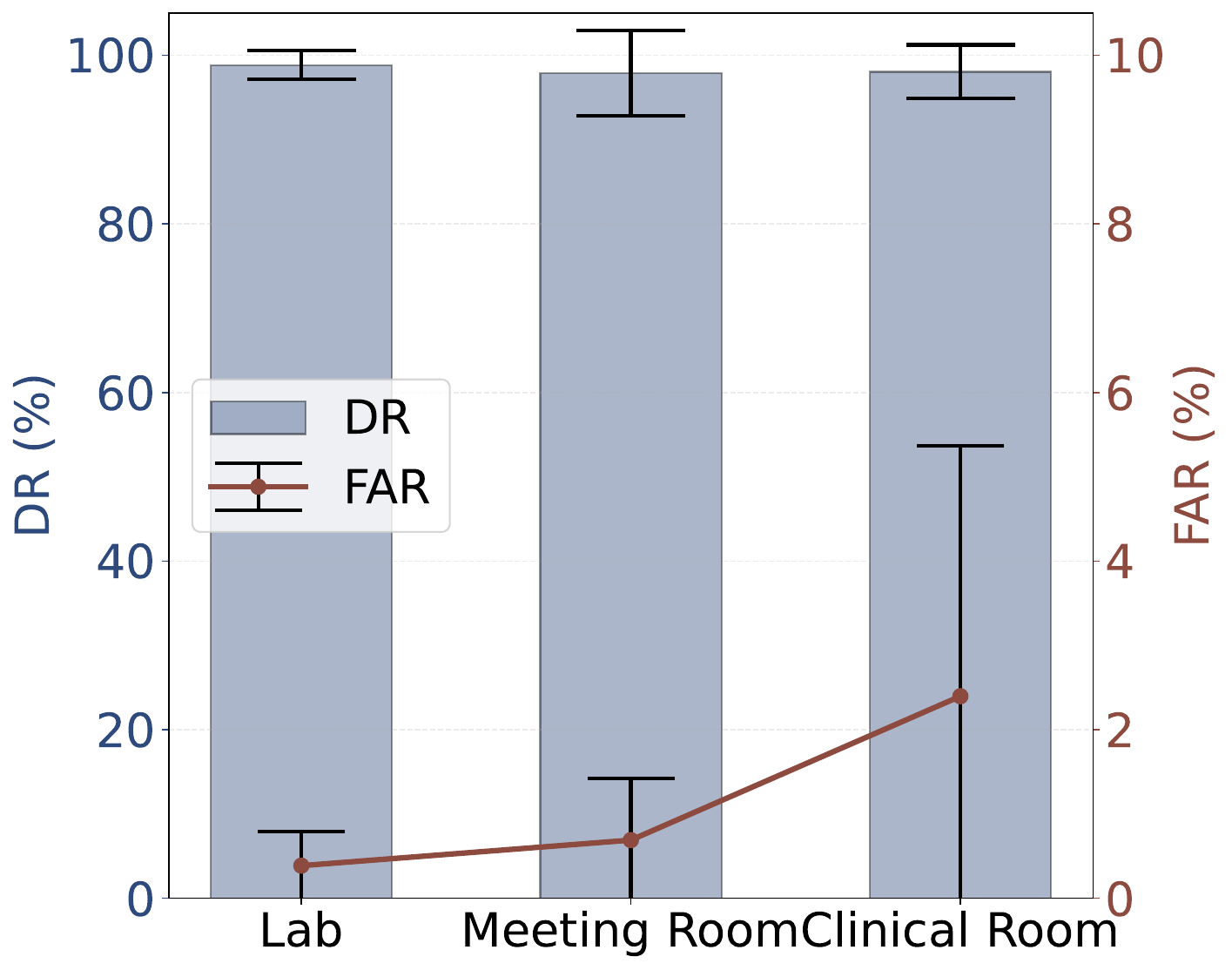}
    \vspace{-1.4\baselineskip}
    \caption{Results across environments.}
    \label{fig:cross env}
\end{minipage}
\begin{minipage}[t]{0.42\textwidth}
\vspace{-2pt}
    \centering
    \includegraphics[width=0.99\textwidth]{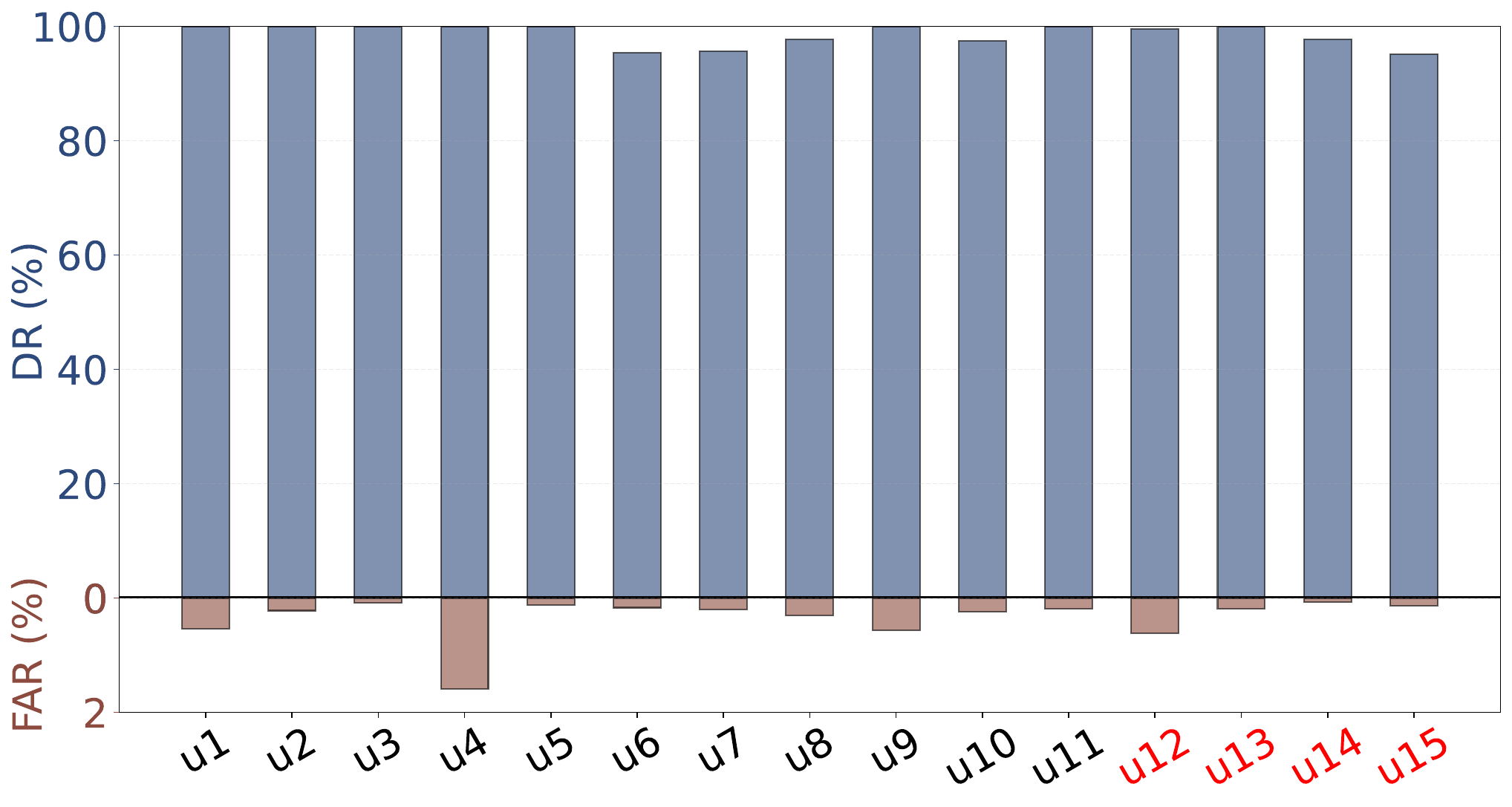}
    \vspace{-1.9\baselineskip}
    \caption{Cross user performance.}
    \label{fig:cross user}
    \end{minipage}

\end{figure*}

4) SeqDet-CNN: 
Many existing thermal-array-based fall detection methods adopt detection- or foreground-guided pipelines to isolate human regions before classification \cite{zhong2020multi, yu2020fall, chen2015fall, hayashida2017use, liu2020fall}.
Following this paradigm, we treat fall detection as a sequence classification task with detection-guided foreground extraction rather than a pure classification problem.
Specifically, \textit{TA CenterDet} is used to localize human regions in each frame and generate masked temperature map sequences by removing background regions, which are then classified using the same convolutional neural networks as the SeqCNN baseline.
This detection-guided masking captures coarse spatio-temporal motion patterns but does not explicitly model pose structure, long-term temporal dynamics, or biomechanical balance information.
It therefore represents a detection-guided sequence classification paradigm that relies on foreground localization rather than physically grounded balance representations.

\subsection{Performance of \sysname}

\head{Overall Performance}
We evaluate the overall performance of \sysname and compare it with representative baselines.
As shown in \fig \ref{fig:overall_performance_distance_f1}, our \sysname achieves a DR (TPR) of 98.26\% with a FAR of 0.65\%, demonstrating the strongest fall-event detection capability while maintaining an ultra-low false alarm level among all compared methods.
In comparison, the statistical EMD-based method yields a FAR of 5.07\% but suffers from an extremely low DR of only 9.49\%, indicating its limited capability in distinguishing fall events from normal activities.
The detection-driven YOLO-CBAM baseline attains a FAR of 2.07\% and a DR of 17.96\%, showing improved performance over EMD but still failing to reliably capture fall-related patterns.
For sequence-based deep learning baselines, Seq-CNN and SeqDet-CNN achieve DR/FAR of 89.19\%/0.23\% and 93.44\%/0.46\%, respectively. 
The inferior performance of these baselines can be attributed to their lack of explicit balance information modeling.
Specifically, EMD relies on global statistical distributions, YOLO-CBAM focuses on appearance-level detection cues, while Seq-CNN and SeqDet-CNN utilize raw or masked temperature map sequences as input to the network.
Without physically grounded representations of human balance, these approaches are more susceptible to environmental variations and complex motion patterns.
In addition, the size of the circles in the figure encodes the standard deviation of DR and FAR across different test settings, where larger circles indicate higher variance and thus lower stability.
In contrast, \sysname exhibits the smallest circle, suggesting that it achieves the most stable and robust performance among all compared methods.

\head{Impact of Environment}
\sysname demonstrates robust performance across various environments. As shown in \fig\ref{fig:cross env}, \sysname is trained with data from a lab environment and a meeting room environment (\fig \ref{fig:experimental_setup}(a) and (b)) and then tested in the Clinical Simulation Room scenario (\fig\ref{fig:experimental_setup}(c)).
\sysname maintains consistent performance, with DR and FAR variations of less than 2\% across all tested environments.

\begin{figure*}[t]

 \begin{minipage}[t]{0.33\textwidth}
 \vspace{0pt}
    \centering 
    \includegraphics[width=0.99\textwidth]{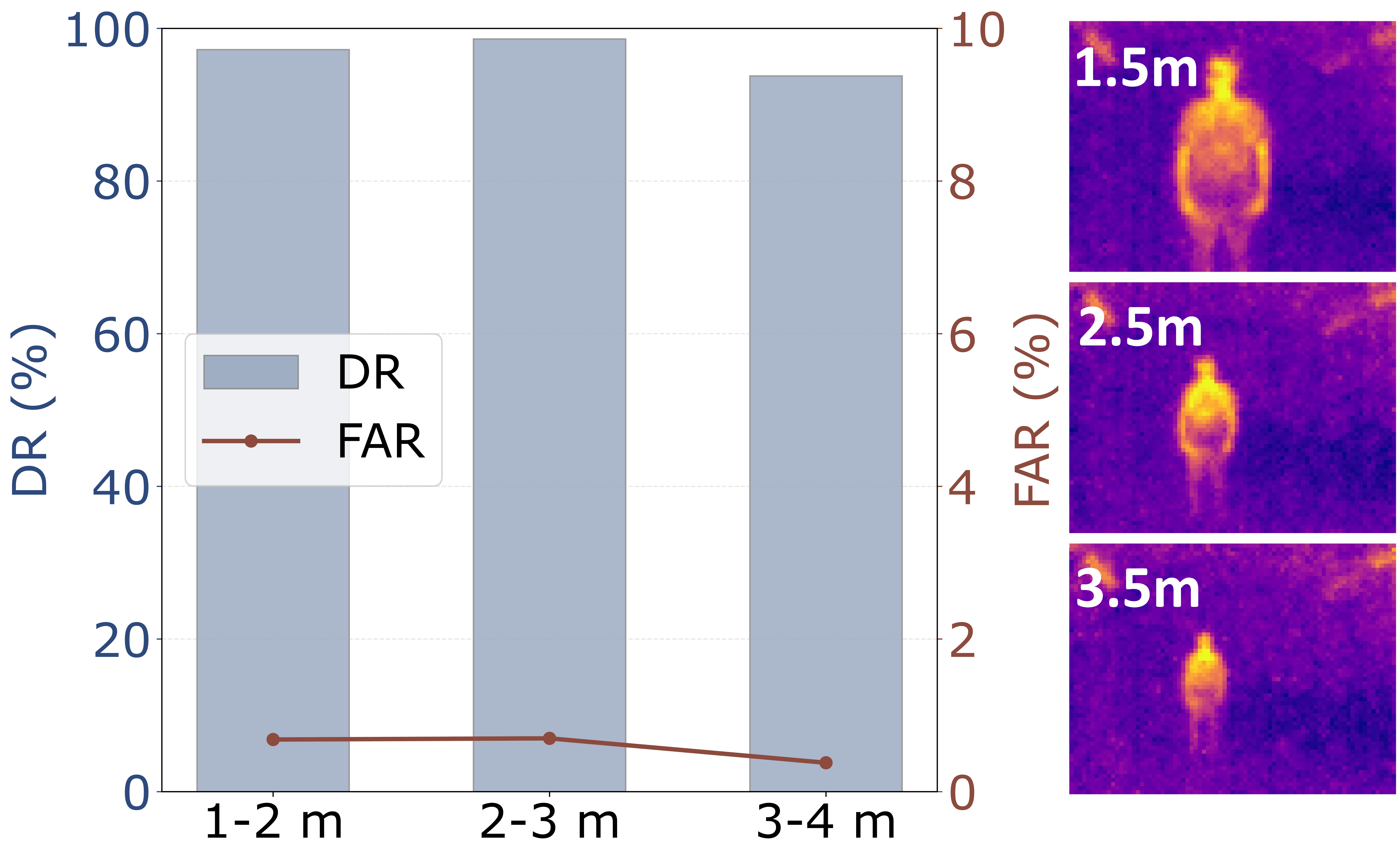}
    \vspace{-1.5\baselineskip}
    \caption{Results across distance.}
    \label{fig:cross distance}
\end{minipage}
  \begin{minipage}[t]{0.33\textwidth}
  \vspace{0pt}
    \centering 
    \includegraphics[width=0.99\textwidth]{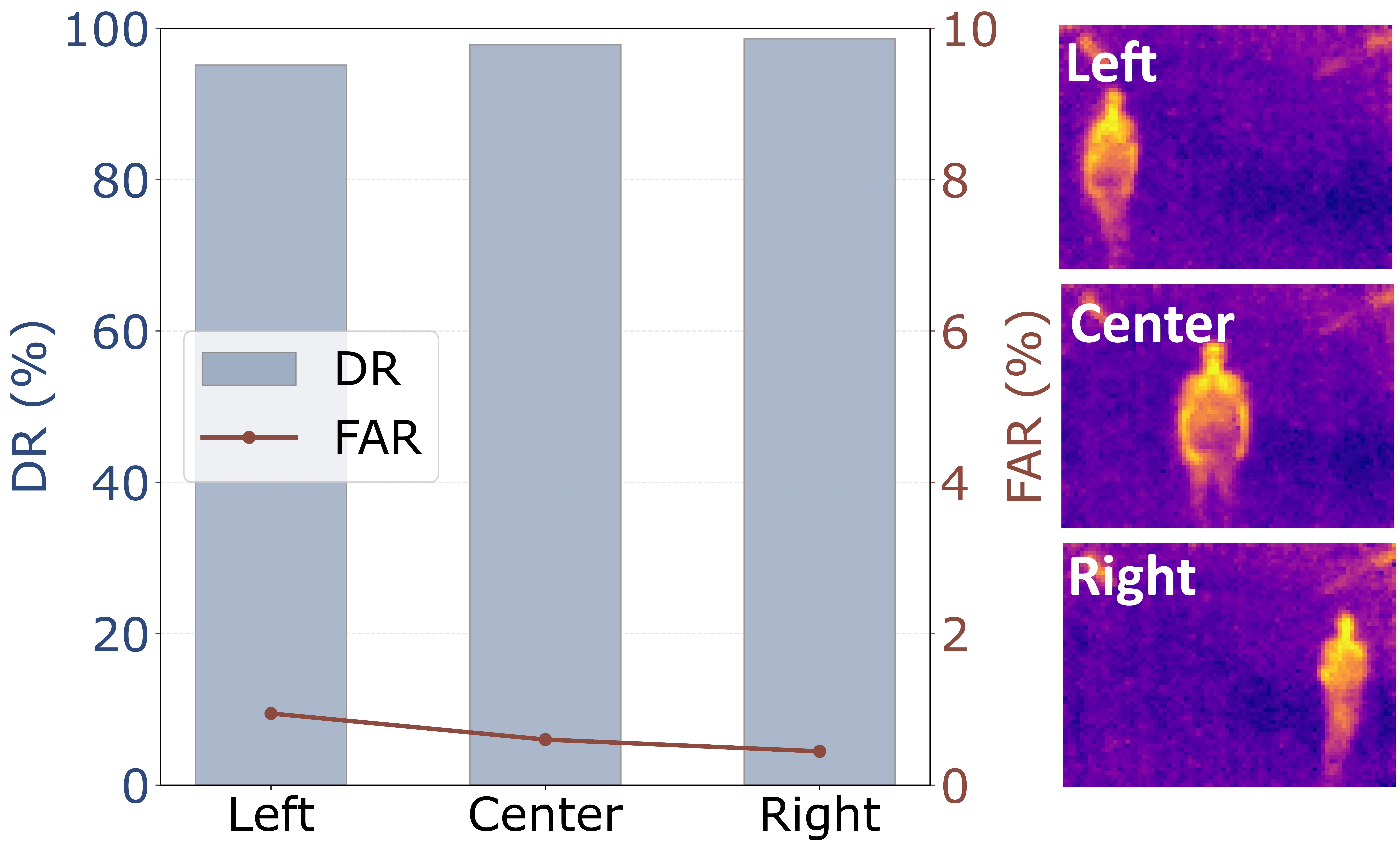}
    \vspace{-1.5\baselineskip}
    \caption{Results across direction.}
    \label{fig:cross angle}
\end{minipage}
\begin{minipage}[t]{0.33\textwidth}
\vspace{0pt}
    \centering
    \includegraphics[width=0.99\textwidth]{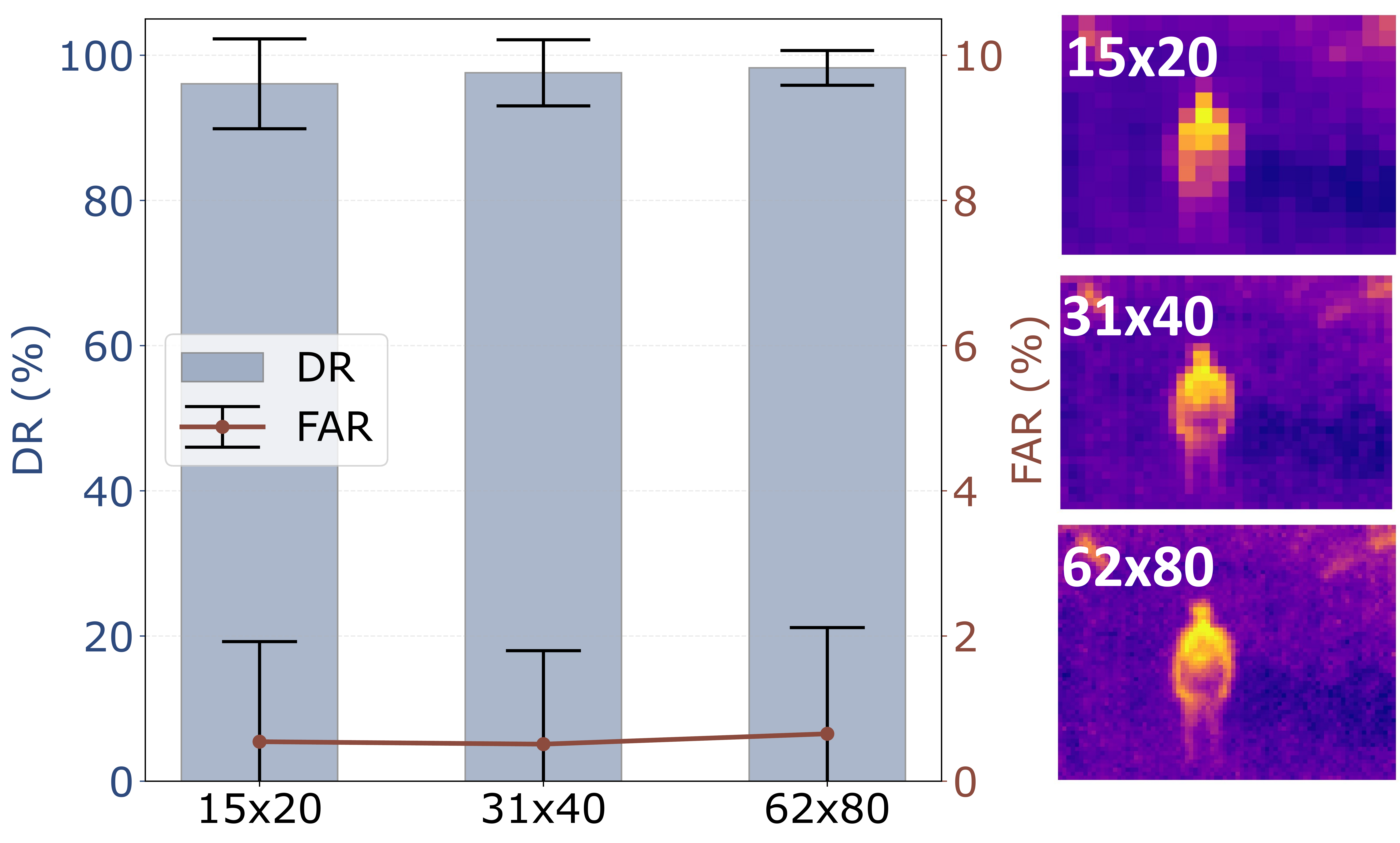}
    \vspace{-1.4\baselineskip}
    \caption{Cross resolution performance.}
    \label{fig:cross resolution}
    \end{minipage}

\end{figure*}

\begin{figure*}[t]
\begin{minipage}[t]{0.60\textwidth}
\vspace{-0pt}
    \centering
    \includegraphics[width=0.99\textwidth]{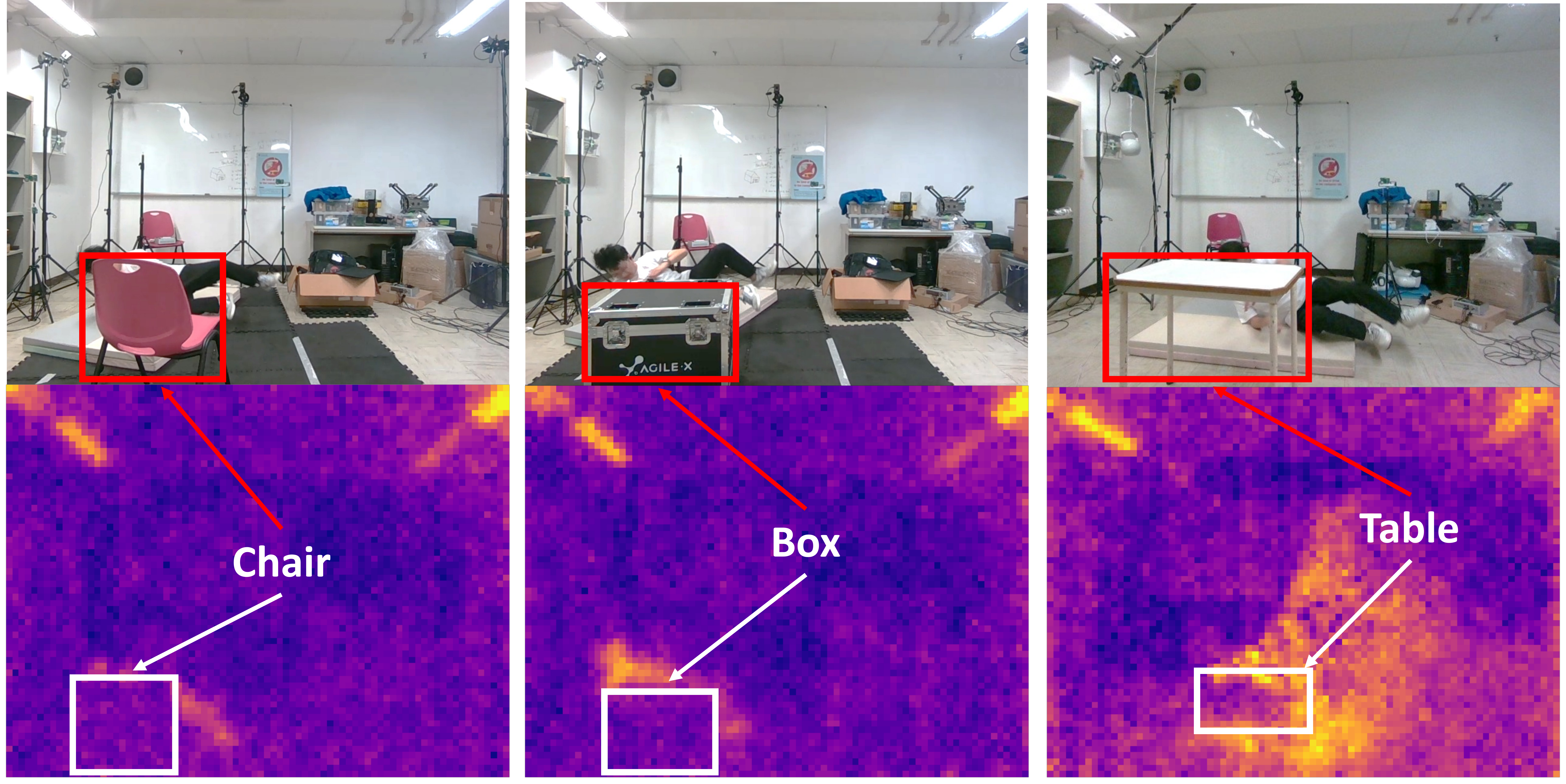}
    \vspace{-1.0\baselineskip}
    \caption{Occlusion of different objects and the impact of object occlusion on temperature map}
    \label{fig:impact_of_occlusion}
    \end{minipage}
\hfill
\begin{minipage}[t]{0.39\textwidth}
\vspace{0pt}
    \centering
    \includegraphics[width=0.99\textwidth]{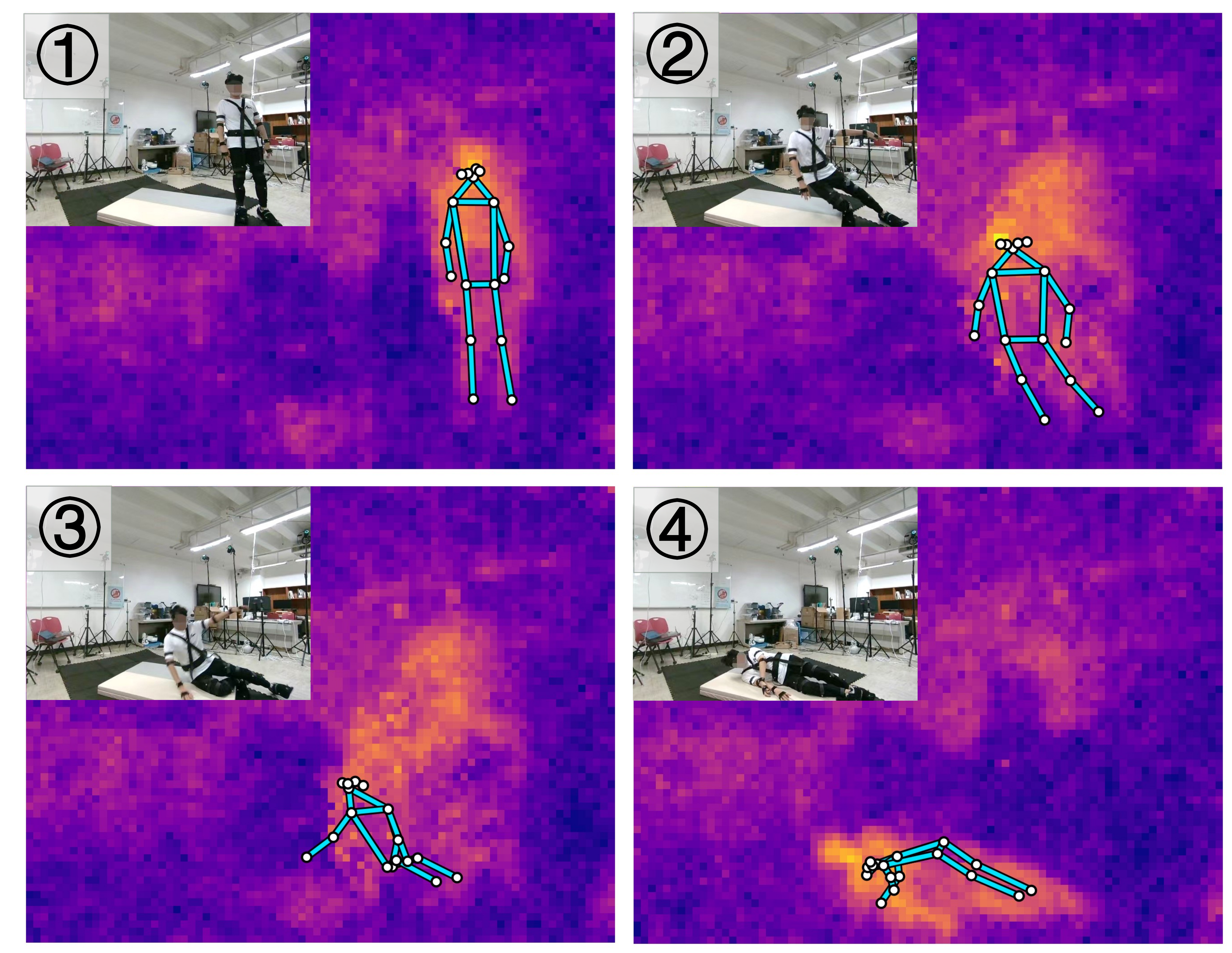}
    \vspace{-2\baselineskip}
    \caption{The visualization of \sysname’s predicted pose during a fall event.}
    \label{fig:skeleton_visual}
    \end{minipage}

\end{figure*}

\head{Impact of Different People} For a practical fall detection system, robust performance across users is crucial. Hence, we assess \sysname's performance using data from 15 participants in the laboratory scenario, labeled U1--U15, by training on a subset of the data from U1--U11 and testing on the rest. As shown in \fig\ref{fig:cross user}, \sysname maintains consistently high fall detection performance. Most user splits achieve DR above 97--100\% while keeping FAR below 2\% (often much lower), demonstrating strong generalization to unseen users. A few splits show slightly lower DR (around 95\%), but the overall trend remains stable, indicating that user-specific motion variations have only a minor effect on the system’s robustness.

\head{Impact of Different Distances} We further evaluate the impact of the distance between the user and the sensor. We group the validation predictions into three distance ranges (1--2 m, 2--3 m, and 3--4 m) and compute performance within each group. As depicted in \fig \ref{fig:cross distance}, \sysname maintains stable performance for distances within 3 m, achieving a DR of 97.2\% and 98.59\% while keeping FAR around 1\%. When the distance increases beyond 3 m (3--4 m), performance of DR decreases slightly by about 3\%. These results indicate that \sysname is robust to moderate variations in user–sensor distance, with reliable fall detection performance maintained within typical indoor deployment ranges.

\head{Impact of Direction}
We evaluate whether \sysname is sensitive to the subject’s horizontal orientation relative to the sensor. As shown in \fig\ref{fig:cross angle}, we partition each temperature map into three horizontal regions (Left, Center, Right) using a 3:2:3 split, and assign each sample to a region based on where the subject is primarily observed. Despite the distinct appearance and motion patterns introduced by different orientations, \sysname maintains consistently high performance across all three regions, with DR and FAR remaining stable and DR consistently above 95\%. The Left region shows a slightly lower score than Center and Right, but the overall difference is small. These results indicate that \sysname does not rely on a specific facing direction and remains robust under orientation changes.

\begin{figure*}[t]
  \centering
  \begin{minipage}[t]{0.35\textwidth}
  \vspace{2pt}
    \centering
    \resizebox{\textwidth}{!}{%
      \begin{tabular}{lcc}
        \hline
        
        Method & Pixel Error & Depth Error \\ \hline
        w/o MB\&Det & 0.0361 & 0.1532 \\ 
        w/o MB & 0.0355 & 0.156 \\ \hline
        \textbf{\sysname} & \textbf{0.0318} & \textbf{0.1265} \\ \hline
      \end{tabular}
    }
    \vspace{0.5\baselineskip}
    \captionof{table}{Pose estimation performance of \sysname.
    MB denotes the Motion Blur Branch and Det refers to TA CenterDet.
    Pixel and depth errors are reported as mean absolute error (MAE) on normalized image coordinates ($[0,1]$) and in meters, respectively.}

    \label{tab:skeleton_performance}
    
  \end{minipage}
  \begin{minipage}[t]{0.29\textwidth}
  \vspace{0pt}
    \centering 
    \includegraphics[width=0.99\textwidth]{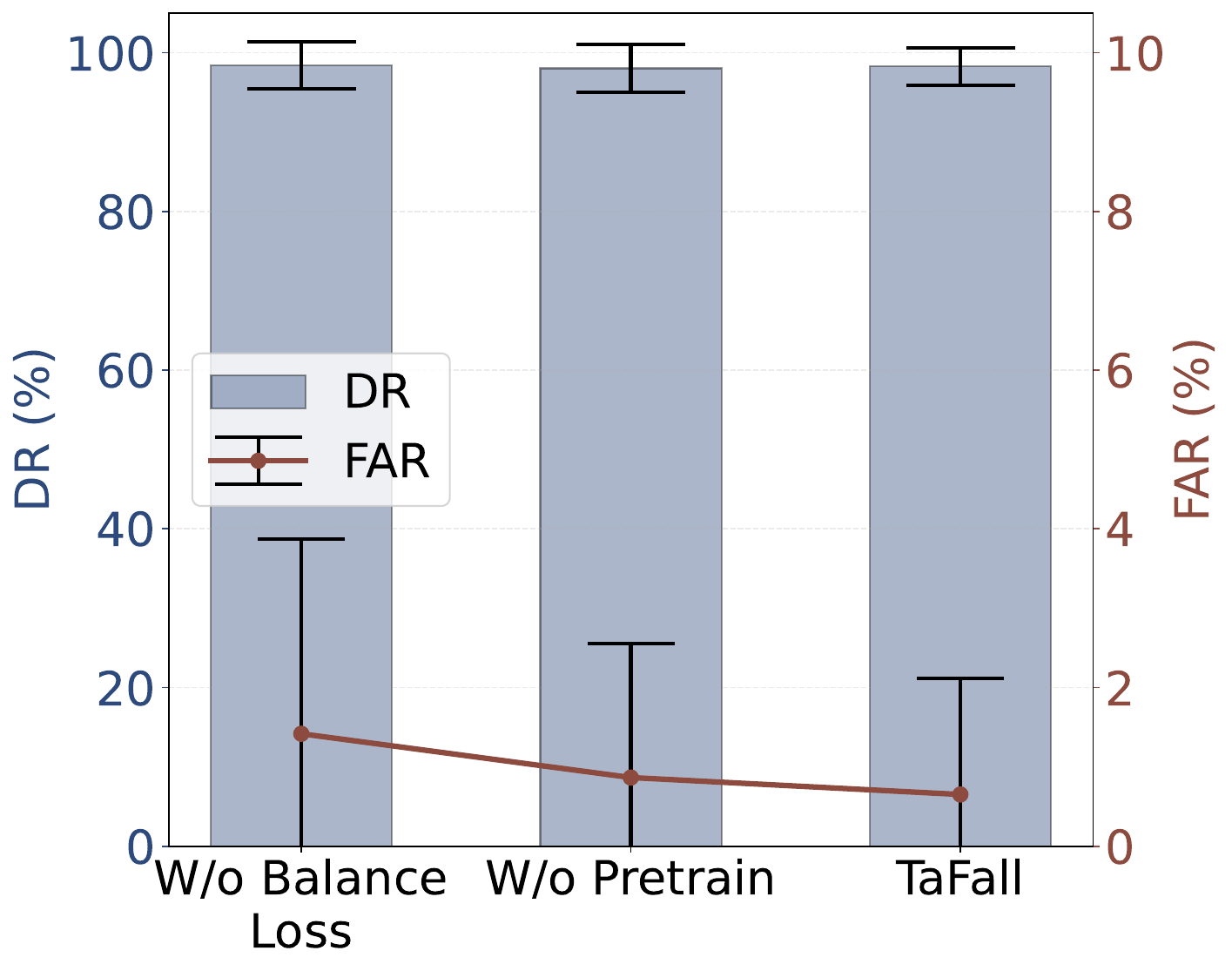}
    \vspace{-1.8\baselineskip}
    \caption{Ablation study results.}
    \label{fig:ablation}
  \end{minipage}
  \hfill
  \begin{minipage}[t]{0.34\textwidth}
  \vspace{-1.1pt}
    \centering 
    \includegraphics[width=0.99\textwidth]{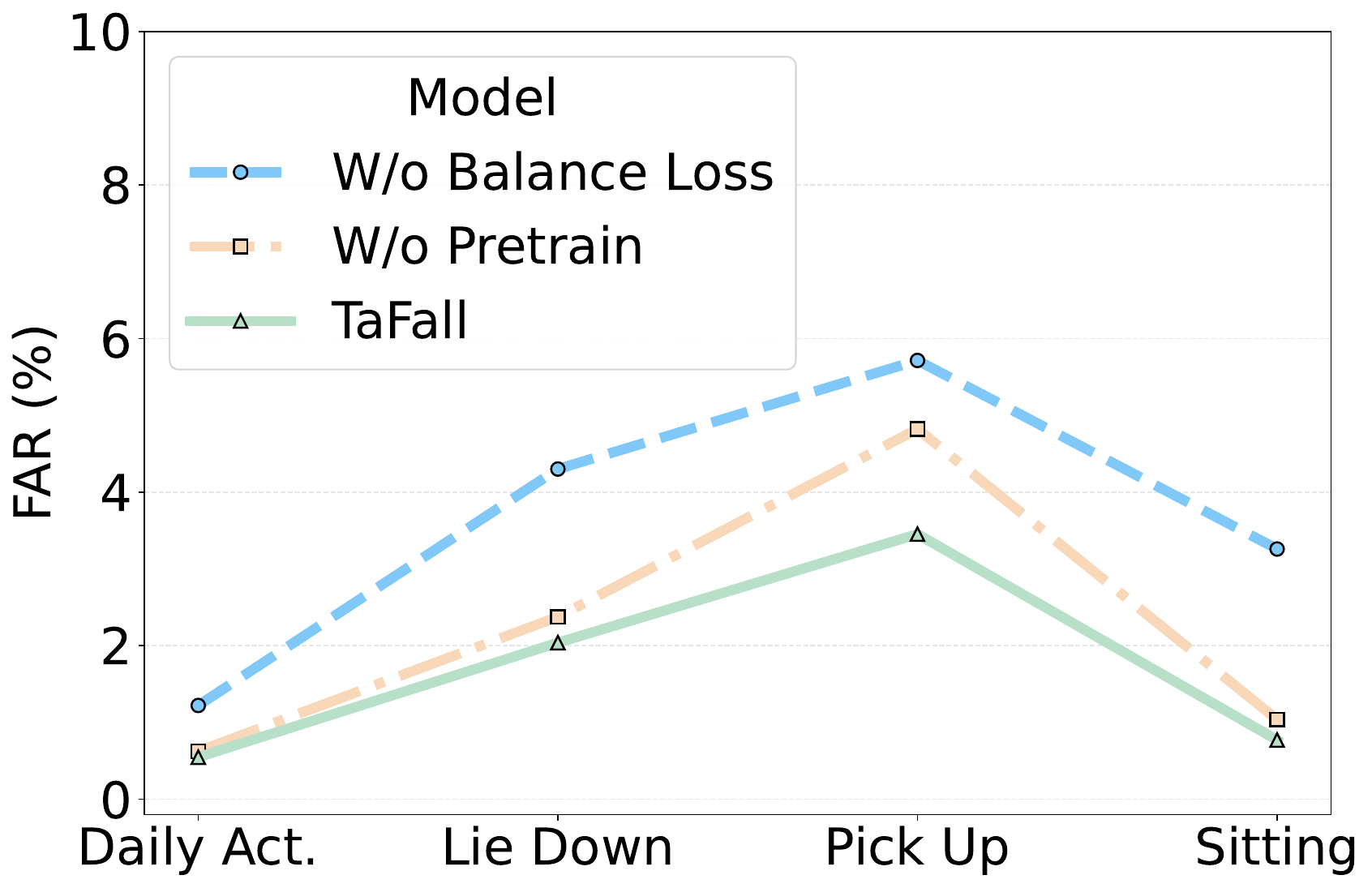}
    \vspace{-1.5\baselineskip}
    \caption{Performance of fall-like activities.}
    \label{fig:fall_like}
  \end{minipage}
  \hfill

\end{figure*}

\head{Impact of Resolution} 
To evaluate the robustness of \sysname across varying spatial resolutions, we downsample the input temperature maps from the original $62\times80$ to $31\times40$ and $15\times20$, respectively. As illustrated in \fig\ref{fig:cross resolution}, the model exhibits remarkable stability despite the reduction in spatial details. Notably, the FAR remains consistently below 1\% across all three resolutions, indicating that \sysname effectively suppresses false alarms even under substantial spatial downsampling. In addition, \sysname achieves a DR of 97.58\% at the $31\times40$ resolution. Even at the coarsest resolution of $15\times20$, the DR degrades only marginally to 96.06\%. These results substantiate the robustness of \sysname to low-resolution inputs, confirming its ability to preserve critical discriminative features even under significant downsampling.

\head{Impact of Occlusion}
As shown in \fig\ref{fig:impact_of_occlusion}, we further evaluate the performance of \sysname under object occlusion, which commonly occurs in indoor environments due to furniture such as chairs, boxes, and tables. Besides,  \fig\ref{fig:impact_of_occlusion} illustrates representative temperature maps captured under chair, box and table occlusions.
When the human body is partially obstructed by surrounding objects, especially during the later stages of a fall when the subject approaches the ground, \sysname may fail to reliably detect the human body, as a large portion of the body is occluded in the thermal observations. 
Despite this limitation, the proposed system remains capable of accurately inferring balance states from incomplete thermal information, as it operates on temporal sequences rather than single frames. 
By leveraging the motion and balance dynamics accumulated over preceding frames, \sysname can maintain reliable balance-state estimation even when instantaneous spatial cues are degraded by occlusion.
Experimental results demonstrate that \sysname successfully detects all 29 fall events under occluded conditions without introducing false alarms. These findings highlight the robustness of our balance-aware representation and its resilience to common occlusions in real-world indoor environments.

\head{Pose and Depth Estimation Performance} 
We evaluate the pose estimation accuracy of \sysname and compare it with two ablated variants that remove the Motion Blur Branch and the detection module.
Since \sysname outputs a 2.5D pose sequence, we report the mean absolute error (MAE) on the image plane (normalized pixel coordinates) and the MAE of depth (in meters).
\fig~\ref{fig:skeleton_visual} further visualizes a representative predicted pose during a fall event, qualitatively illustrating the temporal pose structure recovered from low-resolution thermal observations.
As shown in \tab\ref{tab:skeleton_performance}, \sysname achieves the best performance, reducing the pixel error to 0.0318 and the depth error to 0.1265m.

\begin{figure*}[t]
    \centering
    \includegraphics[width=0.99\textwidth]{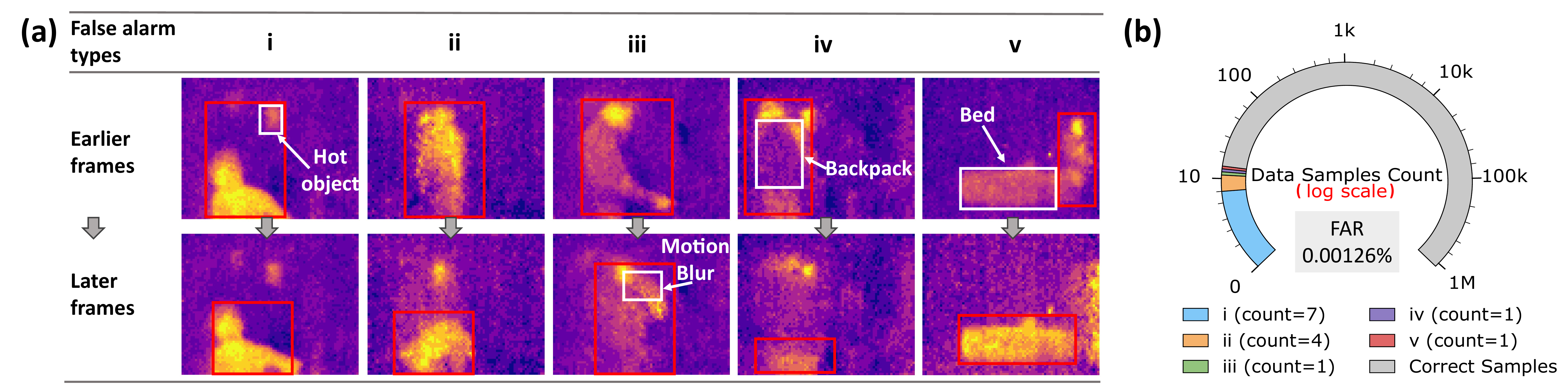}
    \caption{Long-term false alarm case study in real environment. \fig(a) shows five types of false alarms(FA), where the red boxes indicate the bounding boxes predicted by the detection model of \sysname, \fig (b) shows the number of different false alarm types together with the total number of samples.}
    \label{fig:case_long}
\end{figure*}

\subsection{Ablation Study} \sysname comprises two central components: (i) Physically Grounded Balance Representation, and (ii) Pose-Bridged OOV Enhancement. The Balance-Aware Pose Network leverages the balance loss to extract physically meaningful balance features, while the pose-bridged OOV enhancement strategy enhances robustness against OOV behaviors. To understand the contribution of each component, we ablate the balance loss and the pretraining scheme and evaluate their impact on both overall performance and on fall-like activities.
As shown in \fig\ref{fig:ablation}, all three variants exhibit comparable performance, with DR consistently above 98\%. In terms of FAR, \sysname achieves the lowest value (0.65\%), compared with 1.42\% for the variant without Balance Supervision and 0.87\% for the variant without Pose-bridged pretraining. 

However, as shown in \fig\ref{fig:fall_like}, the differences become more pronounced when evaluating fall-like activities, including random daily actions, lying on the ground, picking up objects, and sitting down. To stress the models’ sensitivity to abrupt dynamics, these actions are performed at higher speeds. Under these challenging conditions, \sysname consistently yields the best performance. This suggests that incorporating both the balance loss and pose-bridged pretraining enables the model to attend more closely to balance dynamics rather than being distracted by rapid, non-fall movements.

\section{Case study}

We further conduct two real-world case studies to assess \sysname in practical deployments. 
We first analyze long-term false alarms in residential environments, and then evaluate robustness in a bathroom setting with thermal interference, and occlusions. 
These studies demonstrate reliable fall detection under challenging real-world conditions.

\subsection{Long-Term Case Study in Real Environment}
The false–alarm rate is a critical metric for fall-detection systems; even a seemingly small rate such as $2\%$ can translate into dozens of unnecessary alerts per day in continuous deployments. Consequently, assessing false alarms in long-term real environments is essential for validating practical usability. 

In this case study, we perform a 27-day data collection from three bedrooms and one living room in two elderly care facilities managed by a hospital, as shown in \fig\ref{fig:experimental_setup}(c). 
We evaluate \sysname across these real-world deployment scenarios. To suppress frames without valid human presence, we apply a detection-confidence threshold of $0.5$ and further require the fall-classification probability to exceed a threshold of 95\%. We use the same thresholding strategy in the following bathroom case study.

\begin{figure}[t]
    \centering
    \includegraphics[width=0.99\linewidth]{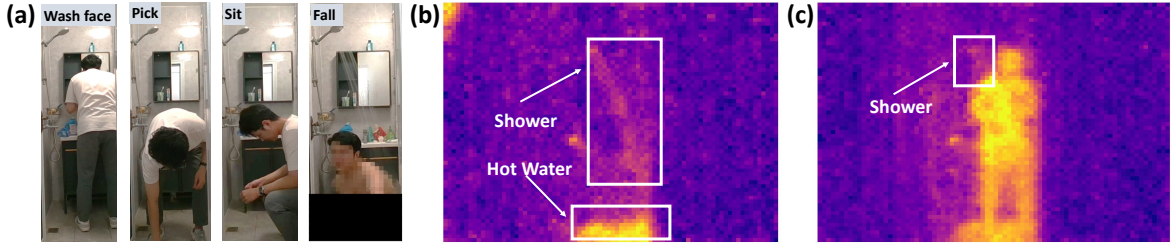}
    \caption{Bathroom case study. (a) Examples of collected daily activities and falls in bathroom environment. (b) False alarm example in a bathroom environment caused by thermal accumulation on the floor when the subject is occluded by the wall. (c) Temperature map during showering in bathroom, where thermal reflections from falling water do not interfere with pose estimation or fall detection.}
    \label{fig:bathroom}
\end{figure}

Under these settings, the overall false alarm rate achieved by \sysname is 0.00126\%. 
The extremely low false alarm rate renders \sysname a promising solution for long-term, real-world adoption. 
To better understand the sources of false alarms, we visualize representative cases, which reveal five distinct categories of causes. 
The first category involves the influence of thermal objects in the room, as shown in \fig \ref{fig:case_long}(a)i. These objects can negatively affect the stability of the detection. This type of false alarm can be mitigated by removing static hot objects from the room.
The second category corresponds to fall-like activities, such as picking up objects, as depicted in \fig \ref{fig:case_long}(a)ii. These activities may be mistakenly classified as falls due to their similar motion patterns.
The third category is also influenced by background thermal objects as shown in \fig \ref{fig:case_long}(a)iii. The key difference here is that in this case, the human body initially overlaps with a hot object and then moves quickly. Once the human body moves away from the hot object, the motion blur overlaps with the thermal object, leading the model to estimate a faster movement of the pose, resulting in a false alarm classification.
The last two categories are particularly interesting. As shown in \fig \ref{fig:case_long}(a)iv, the subject, wearing a backpack, turned their back to the sensor, causing the heat from their torso to be blocked by the backpack. When the subject moved quickly, the model mistakenly identified the lower half of the body as the entire human body, leading to a false alarm.
In \fig \ref{fig:case_long}(a)v, we present another false alarm scenario where the subject stood up and stepped out of the sensor's FoV. When the subject exited the screen, the model mistakenly detected the warm part of the bed as the human body, causing another false alarm. The number of false alarms in each of the five categories is shown in \fig \ref{fig:case_long}(b).

\subsection{Bathroom Fall Detection Case Study}
To assess the \sysname’s performance in privacy-sensitive and safety-critical environments, as shown in \fig \ref{fig:bathroom}, we conducted a real-world case study in a bathroom, where the majority of falls occur. As illustrated in \fig~\ref{fig:bathroom}(a), we collected a set of natural daily activities in everyday routine, including washing the face, sitting down, picking up items, etc. We further performed 25 falls during showering with diverse falling directions and motion patterns. All simulated falls were successfully detected, demonstrating that our system maintains reliable fall detection capability even under challenging indoor moisture and temperature conditions.

To verify the necessity of applying thresholds, we additionally evaluated the system without using any thresholds. In this setting, we observed one false alarm in a bathroom scenario. As shown in \fig \ref{fig:bathroom} (b), when the subject becomes occluded by the bathroom wall, the thermal array immediately captures the heat accumulation from hot water on the floor (white box). This short-term temperature change causes an abrupt thermal variation, leading to incorrect motion estimation and ultimately triggering a false alarm. With the default thresholds (0.95 for fall detection confidence and 0.5 for human detection), this false alarm is suppressed.

In contrast, showering presents minimal disturbance to the system. \fig~\ref{fig:bathroom}(c) shows that the thermal reflections caused by falling water(in the white box) exhibit much weaker spatial structure and therefore do not interfere with pose estimation or fall detection. Moreover, the coarse spatial resolution of the thermal array inherently preserves user privacy in sensitive bathroom activities, while still allowing reliable balance assessment.

\section{Discussions and Future Work}
\label{sec:discussions}

\head{Edge Deployment}
Currently, our system operates on a GPU-equipped computer in real-time at 20 Hz. 
Given the privacy-preserving advantage of thermal array sensing, we can follow the system structure in \sysname that streaming the thermal array data to computing center for further processing.
While, for the edge deployment, since the current \sysname model is transformer-based architecture, the computational demands present hinder the real-time processing.
To address this limitation, future work will explore model compression and knowledge distillation techniques, using \sysname as a teacher model to train a lightweight student network (mini-\sysname) tailored for edge inference.
This strategy aims to preserve the core balance-aware capabilities of the original model while enabling efficient, low-latency, and scalable deployment on edge devices.

\head{Thermal Array vs. mmWave Radar}
Recently, mmWave radar has attracted increasing attention for fall detection \cite{meng2025gr, zhang2023lt}.
However, compared with thermal arrays, mmWave radar typically exhibits lower spatial resolution due to the limited number of antennas.
In contrast, thermal arrays provide higher spatial resolution while also inherently preserving privacy, as they capture textureless temperature maps rather than appearance-level visual information \cite{zhang2024tadar}.
Moreover, mmWave radar requires active signal transmission, resulting in relatively high power consumption, typically exceeding 2 W \cite{zhang2025tapor}.
By comparison, the thermal array used in \sysname consumes only 40 mW.
From a cost perspective, mmWave radar systems are also substantially more expensive than thermal arrays (\eg, \$358.80 for IWR1843BOOST vs. \$18.35 for Meridian MI0802M6S).

\head{Multi-Sensor Fusion} 
While our prototype uses a single thermal array sensor, multi-sensor and multi-view fusion could further improve the robustness under severe occlusion and enable larger coverage. Thermal arrays, as a fully passive modality that does not transmit any signals, exhibit unique advantages for multi-sensor fusion, since the sensors do not introduce mutual interference while other modalities like mmWave and Wi-Fi do.

\head{Toward Environment-Aware Balance Sensing}
Our current \sysname primarily relies on thermal array sensing as the input modality and performs fall detection based on predicted human poses.
While thermal arrays are inherently human-centric, they exhibit limited capability in perceiving environmental context.
As a result, in non-standard postures such as sitting on a chair or lying on a bed, the estimated human pose is inherently “suspended” in space across sensing modalities.
The key difference lies not in pose estimation itself, but in the ability to perceive environmental support.
While other modalities can further capture surrounding structures such as chairs or beds, thermal arrays lack explicit environmental awareness, making it difficult to distinguish genuine balance loss from posture transitions supported by external objects.
To address this limitation, future work will explore multi-modal sensing by integrating thermal arrays with complementary modalities, such as mmWave radar and direct Time-of-Flight (DToF) sensors \cite{li2025facial,liu2025privacy}.
Such cross-modal fusion can enhance environmental awareness and enable more robust interpretation of human posture under diverse environmental conditions, thereby improving the adaptability of balance-informed fall detection across complex real-world scenarios.

\head{Pre-Impact Fall Detection}
Pre-impact fall detection aims to identify an imminent fall during the falling phase, before body-ground impact occurs \cite{yu2021large}.
Most existing fall detection systems primarily focus on recognizing fall events after they have occurred, often relying on abrupt changes in body height, acceleration, or posture that are most salient at or after ground contact \cite{hu2021defall, zhang2023lt, meng2025gr}. While such cues could potentially support early warning, they are commonly formulated within discrete event-based detection frameworks rather than as models of fall progression. However, \sysname introduce the body balance dynamic for fall detection, and formulate the fall event as a progression of the balance degradation. 
Hence, it provides the potential to support earlier fall-risk inference prior to impact by leveraging trends in balance dynamics rather than relying solely on post-impact observations.

\head{Beyond Fall Detection}
An exciting future direction is to extend our balance-informed formulation from binary fall detection to richer fall analytics, \eg, pre-impact fall detection, fall-risk assessment, fall prevention, post-fall state monitoring, \etc. Our future work will also explore large-scale real-world deployment of \sysname. 
This will broaden the clinical value of passive thermal sensing for elderly care in everyday living environments.

\section{Related Works}
\label{sec:related_works}
In this section, we review related work on fall detection systems across sensing modalities, including wearable, camera, RF, and thermal approaches. Detailed comparisons are provided below.

\head{Wearable-based}
Most works in fall detection using wearable devices are based on wearable IMUs \cite{picerno2021wearable, hu2016pre, nait2018deep, palmerini2020accelerometer, bagala2012evaluation, bourke2016fall}. These works generally relay on either threshold based algorithms or ML approaches \cite{picerno2021wearable}. For example, Bagala\etal \cite{bagala2012evaluation} benchmarks thirteen fall detection algorithms using IMU sensors based on threshold on real-world falls from high-risk patients, offering the first systematic evaluation beyond simulated scenarios. On the other hand Alan \etal \cite{bourke2016fall} evaluates machine learning-based fall detection algorithms using features extracted from the FARSEEING real-world fall dataset.
Similarly, Palmerini \etal \cite{palmerini2020accelerometer} leverages the largest set of real-world fall data to develop machine learning algorithms based on a multiphase fall model, achieving improved performance and offering insights into designing practical, real-life fall detection systems.
However, wearable-based systems require consistent usage and regular charging, but these expectations are often difficult to meet for elderly users. Due to discomfort and inconvenience, such devices are frequently underutilized, and cognitive decline or forgetfulness may lead to them being forgotten or left uncharged, compromising system reliability.

\head{Camera-based}
Vision data from cameras is increasingly utilized for fall detection due to its numerous advantages over wearable devices These advantages include the ability to detect multiple events simultaneously, suitability for various subjects, environments, and tasks, as well as ease of installation and visual verification of data \cite{nizam2020classification}. The CV based method can be devided into 3 classes: pose Estimation \cite{inturi2023novel, raza2023logrf, zahan2022sdfa, li2022fall, saini2019kinect, pranavan2023fall}, Object Detection \cite{fernando2021computer,zheng2022realization, killian2021fall}and other features \cite{patel2024ai, wang2023fall, romaissa2020fall}. For example, Beddiar\etal \cite{romaissa2020fall} proposes a fall detection method based on human body geometry extracted from video frames, using angular and distance features between the head and hip to train an SVM and LSTM. On the other hand, Intrui\etal \cite{inturi2023novel} utilize AlphaPose to extract pose keypoints, employing a CNN-LSTM framework to capture spatial and temporal patterns and demonstrates strong performance on the UP-FALL dataset compared to OpenPose-based methods. What's more, Chen \etal \cite{chen2021fall}proposes a fall detection method that combines pose estimation with an auxiliary YOLOv5-based object detection approach, enabling accurate classification and localization of falls in surveillance videos. However, camera-based methods face several challenges, including significant privacy concerns, sensitivity to lighting conditions, and substantial computational resource requirements.

\head{RF-based} 
Wireless and acoustic signals are exploited as more privacy-friendly alternatives for fall detection. Wireless signals used for fall dection including WiFi \cite{palipana2018falldefi, wang2016rt, wang2016wifall, hu2021defall, ji2022sifall}, Radar \cite{zhang2023lt,liu2011automatic, liu2012doppler, shrestha2017feature, li2017multisensor,sadreazami2019fall, su2022hybrid}, Ultrasound \cite{popescu2008acoustic, khan2015unsupervised, adnan2018fall,lian2021fall} and LoRa \cite{zhang2023lofall}. For instance, Hu \etal \cite{hu2021defall} proposed DeFall, a WiFi-based passive FD system that is independent of the environment and free of prior training in new environments. Besides, Zhang \etal \cite{zhang2023lt} proposed LT-Fall, a mmWave-based system that detects life-threatening falls by combining fall detection with post-fall immobility analysis. Although RF-based methods mitigate the privacy concerns inherent in camera-based approaches, their limited spatial resolution leads to reliance on coarse motion cues, which are easily confounded by fall-like daily activities and environmental dynamics, thereby reducing robustness in real-world deployment.

\head{Thermal-based}
Compared with the above modalities, thermal-based approaches offer an attractive tradeoff between privacy preservation and effective human sensing \cite{zhang2024tadar}
Several recent studies have explored fall detection using thermal array sensor \cite{newaz2025approach, jiang2024fall, naser2022multiple, zhong2020multi, rezaei2021unobtrusive, yu2020fall, adolf2018deep, hayashida2017use, liu2020fall, chen2015fall}. 
For example, Naser \etal \cite{naser2022multiple} proposed a privacy-aware method that replaces raw temperature values with optical flow vectors and applies a Bi-LSTM model for activity classification.
Newaz and Hanada \cite{newaz2025approach} proposed a statistical fall detection approach using low-resolution infrared array sensors, which models thermal signature distributions and distinguishes falls via Earth Mover’s Distance (EMD)–based divergence without relying on learned pose or structural representations.
Jiang \etal \cite{jiang2024fall} treated infrared-array-based fall detection as an object detection problem using a modified YOLOv5 network with the Convolutional Block Attention Module (CBAM), enabling attention-enhanced fall pattern localization from thermal maps.
However, many existing methods treat thermal array measurements simply as low-resolution image sequences and apply conventional vision pipelines, without fully accounting for sensing-specific characteristics such as extremely low spatial resolution, motion-induced blur, and relatively high sensor noise.
In addition, most prior studies are evaluated only on short-term or laboratory datasets, with limited evidence from long-term real-world deployments.

In contrast, we propose \sysname, to the best of our knowledge, the first thermal array based, balance-informed fall detection system designed and validated for real-world indoor environments. Rather than modeling falls purely as appearance or motion pattern classification problems, \sysname formulates a fall as a progressive degradation of human balance grounded in biomechanics. By elevating low-resolution thermal observations to pose dynamics while explicitly incorporating thermal sensing characteristics, and further mapping them to physically interpretable balance representations, our approach achieves high detection accuracy together with consistently low false alarm rates in long-term real-world deployments.

\section{Conclusions}
\label{sec:conclusion}

We present \sysname, the first balance-informed fall detection system using low-cost thermal arrays. 
Rather than focusing on coarse motion information such as height changes or velocity patterns, \sysname models a fall as a dynamic progression of balance degradation and detects falls through pose-driven, biomechanically grounded balance dynamics. 
\sysname achieves so via three novel designs: appearance-motion fusion that reconstructs 2.5D pose sequences from low-resolution, motion-blurred temperature maps, physically grounded balance-aware learning based on the Signed Margin of Balance, and pose-bridged pretraining to improve robustness to diverse behaviors and viewpoint variants. 
Across large-scale experiments with over 3,000 fall samples from 35 participants and long-term deployment spanning 27 days in four elderly homes, \sysname achieves high detection accuracy with ultra-low false alarms, promising a practical fall detection solution for real-world adoption. 

\bibliographystyle{ACM-Reference-Format}
\bibliography{refs}

\end{document}